\documentclass[acmlarge]{acmart}

\usepackage{amsmath}
\usepackage{amsfonts}
\usepackage{algorithmic}
\usepackage{graphicx}
\usepackage{textcomp}
\usepackage{xcolor}

\usepackage[caption=false]{subfig}
\usepackage{svg}
\usepackage{url}
\usepackage{multicol}
\usepackage{multirow}
\usepackage{color}
\usepackage{hhline}
\usepackage{pifont}
\usepackage{comment}
\usepackage{cleveref}
\usepackage{tablefootnote}
\usepackage{flushend}

\usepackage{colortbl}
\usepackage[ruled,vlined]{algorithm2e}
\definecolor{mynicegreen}{RGB}{11,102,35}
\definecolor{myniceblue}{rgb}{154,201,219}
\usepackage[group-separator={,}]{siunitx}

\newcommand{\model}{{\textsc{DHC-HGL}}}

\newcommand{\etal}{{\textit{et al.}}}
\newtheorem{definition}{Definition}

 \newcommand{\squishlist}{
	\begin{list}{$\bullet$}
		{ \setlength{\itemsep}{0pt}
			\setlength{\parsep}{3pt}
			\setlength{\topsep}{3pt}
			\setlength{\partopsep}{0pt}
			\setlength{\leftmargin}{1.5em}
			\setlength{\labelwidth}{1em}
			\setlength{\labelsep}{0.5em} } }
	
	\newcommand{\squishlisttwo}{
		\begin{list}{$\bullet$}
			{ \setlength{\itemsep}{0pt}
				\setlength{\parsep}{0pt}
				\setlength{\topsep}{0pt}
				\setlength{\partopsep}{0pt}
				\setlength{\leftmargin}{2em}
				\setlength{\labelwidth}{1.5em}
				\setlength{\labelsep}{0.5em} } }
		
		\newcommand{\squishend}{
	\end{list}}
\AtBeginDocument{%
  \providecommand\BibTeX{{%
    \normalfont B\kern-0.5em{\scshape i\kern-0.25em b}\kern-0.8em\TeX}}}

\setcopyright{acmlicensed}
\acmJournal{IMWUT}
\acmYear{2023} \acmVolume{7} \acmNumber{4} \acmArticle{159} \acmMonth{12} \acmPrice{15.00}\acmDOI{10.1145/3631444}

\begin{document}

\title{Deep Heterogeneous Contrastive Hyper-Graph Learning for In-the-Wild Context-Aware Human Activity Recognition}

\author{Wen Ge}
\authornote{Equal Contribution.}
\email{wge@wpi.edu}
\orcid{0000-0002-9247-1162}
\author{Guanyi Mou}
\orcid{0000-0002-9987-0342}
\authornotemark[1]
\email{gmou@wpi.edu}
\affiliation{%
  \institution{Worcester Polytechnic Institute}
  \streetaddress{100 Institute Road}
  \city{Worcester}
  \state{MA}
  \country{USA}
  \postcode{01605}
}

\author{Emmanuel O. Agu}
\orcid{0000-0002-3361-4952}
\affiliation{%
  \institution{Worcester Polytechnic Institute}
  \streetaddress{100 Institute Road}
  \city{Worcester}
  \country{USA}}
\email{emmanuel@wpi.edu}

\author{Kyumin Lee}
\orcid{0000-0002-9004-1740}
\affiliation{%
  \institution{Worcester Polytechnic Institute}
  \streetaddress{100 Institute Road}
  \city{Worcester}
  \state{MA}
  \country{USA}}
\email{kmlee@wpi.edu}

\renewcommand{\shortauthors}{Ge and Mou et al.}

\begin{abstract}

{Human Activity Recognition (HAR) is a challenging, multi-label classification problem as activities may co-occur and sensor signals corresponding to the same activity may vary in different contexts (e.g., different device placements). This paper proposes a Deep Heterogeneous Contrastive Hyper-Graph Learning (DHC-HGL) framework that captures heterogenous Context-Aware HAR (CA-HAR) hypergraph properties in a message-passing and neighborhood-aggregation fashion. Prior work only explored homogeneous or shallow-node-heterogeneous graphs. DHC-HGL handles heterogeneous CA-HAR data by innovatively 1) Constructing three different types of sub-hypergraphs that are each passed through different custom HyperGraph Convolution (HGC) layers designed to handle edge-heterogeneity and 2) Adopting a contrastive loss function to ensure node-heterogeneity.
In rigorous evaluation on two CA-HAR datasets, DHC-HGL significantly outperformed state-of-the-art baselines by 5.8\% to 16.7\% on Matthews Correlation Coefficient (MCC) and 3.0\% to 8.4\% on Macro F1 scores. UMAP visualizations of learned CA-HAR node embeddings are also presented to enhance model explainability.} Our code is publicly available\footnote{\url{https://github.com/GMouYes/DHC_HGL}} to encourage further research.
\end{abstract}

\ccsdesc[500]{Human-centered computing~Ubiquitous and mobile computing}
\ccsdesc[300]{Computing methodologies~Supervised learning}
\keywords{human activity recognition, heterogeneous graph, hypergraph, graph neural networks}


\maketitle

\begin{figure}[htbp]
    \centering
    \subfloat[$User_1$ \textit{walking} with \textit{phone in bag}]{
        \includegraphics[width=0.305\linewidth]{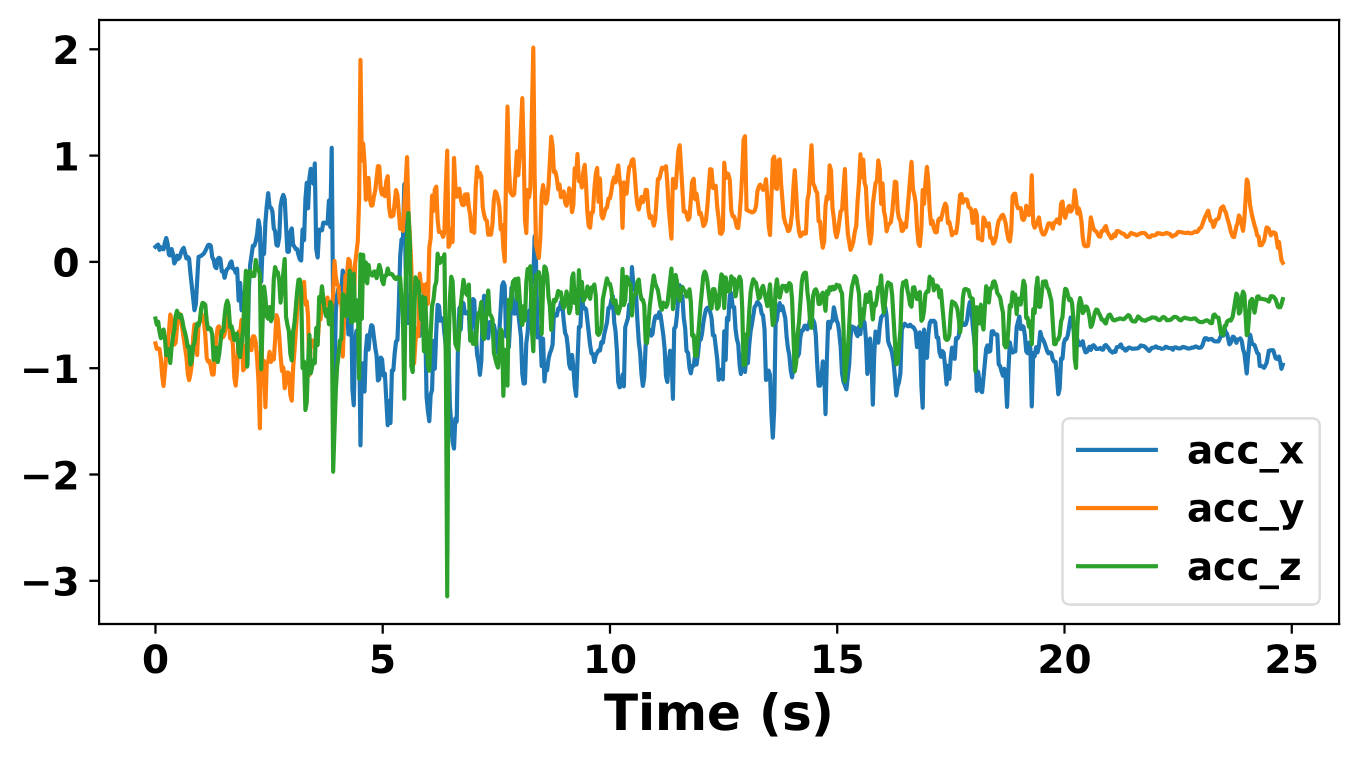}
        \label{fig:u1walkbag}
    } \hspace{-0.5cm}
    \qquad
    \subfloat[$User_1$ \textit{walking} with \textit{phone in hand}]{
        \includegraphics[width=0.305\linewidth]{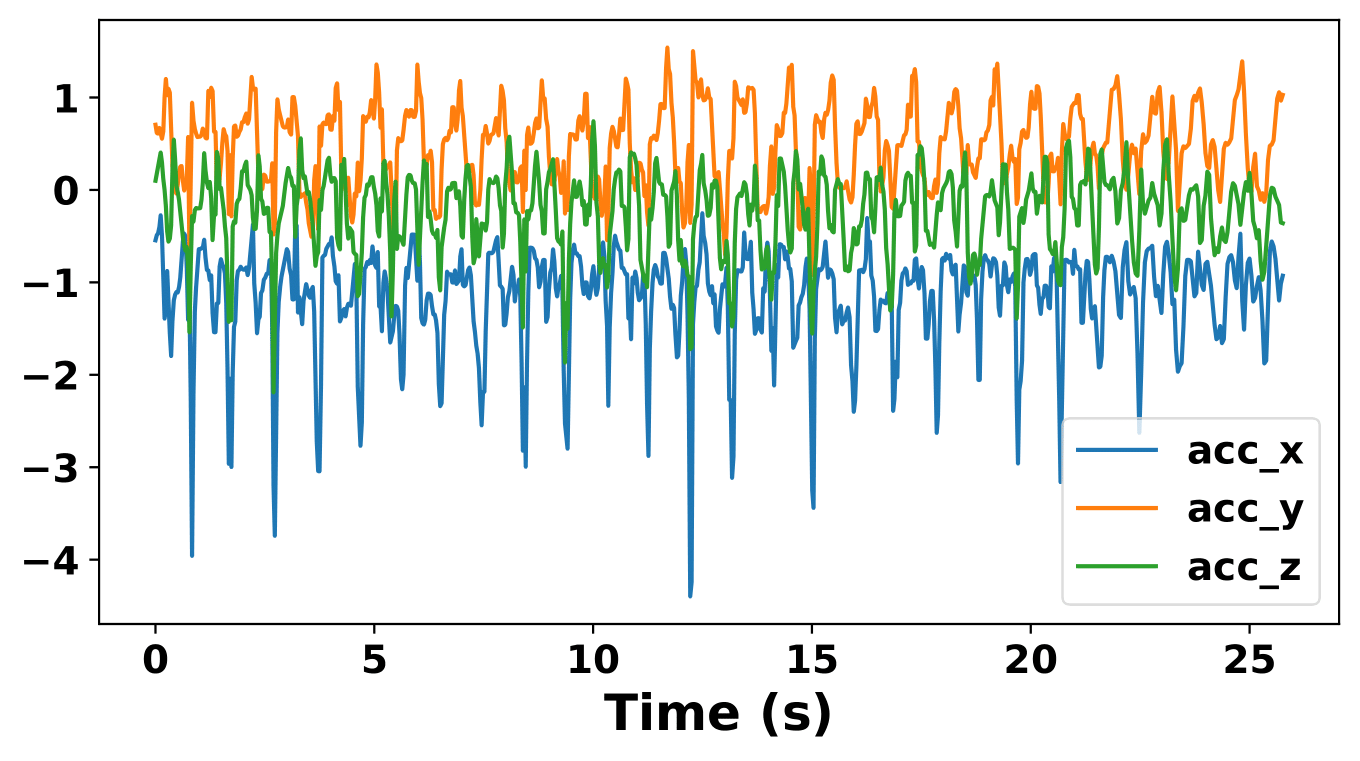}
        \label{fig:u1walkhand}
    }\hspace{-0.5cm}
    \qquad
     \subfloat[$User_2$ \textit{walking} with \textit{phone in hand}]{
        \includegraphics[width=0.305\linewidth]{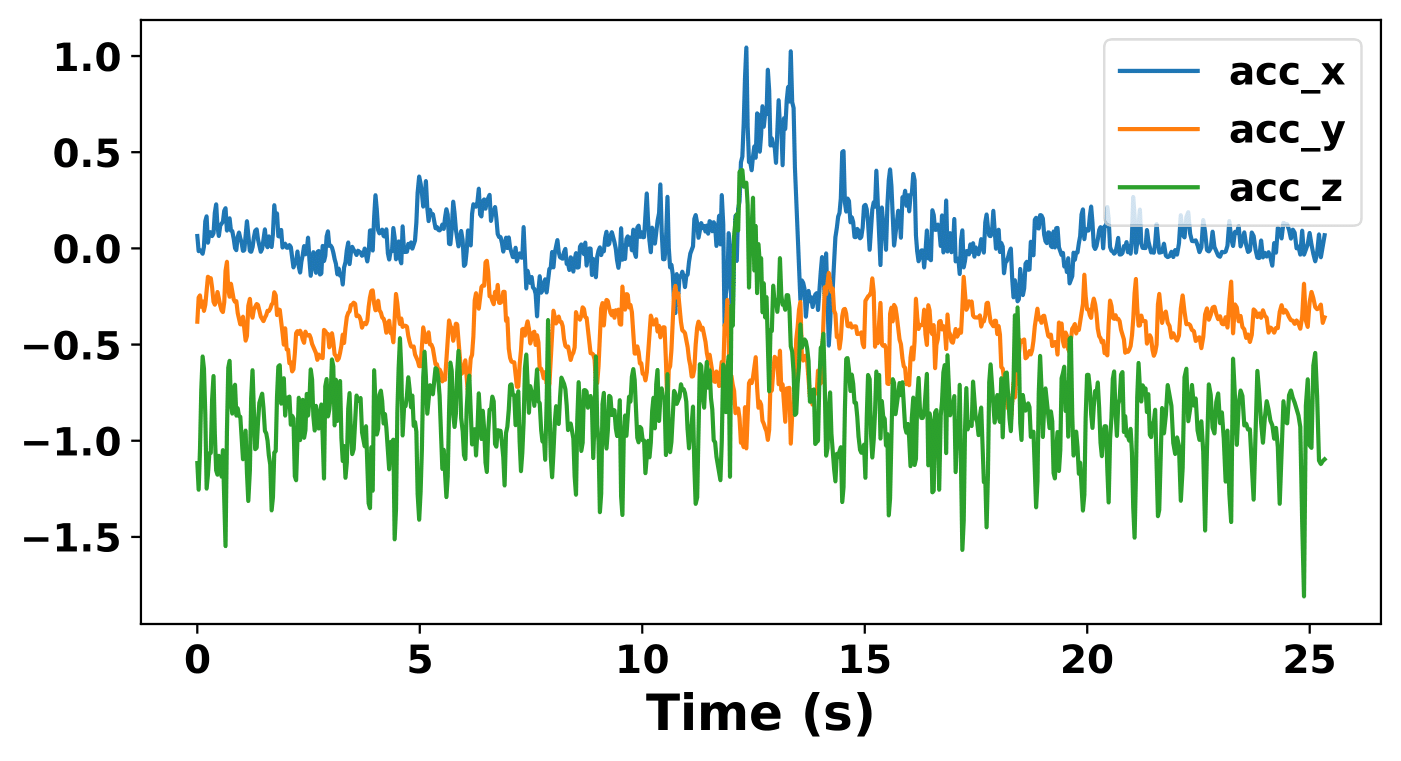}
        \label{fig:h2walkhand}
    }
    \caption{Accelerometer signal corresponding to the Walking activity in various contexts and performers from real-world \textit{Extrasensory} dataset~\cite{vaizman2017recognizing}. Comparing Fig.~\ref{fig:u1walkbag} and Fig.~\ref{fig:u1walkhand}, we observe disparate accelerometer readings of the same activity under different contexts. Meanwhile, as we compare Fig.~\ref{fig:u1walkhand} and Fig.~\ref{fig:h2walkhand}, different users might perform the same activity differently, producing distinct sensor readings, even with the same contextual factors.}
    \label{fig:contextNuser}
\end{figure}

\section{Introduction}
\label{sec:intro}

\begin{figure}[t]
\centering
    \includegraphics[width=.8\linewidth]{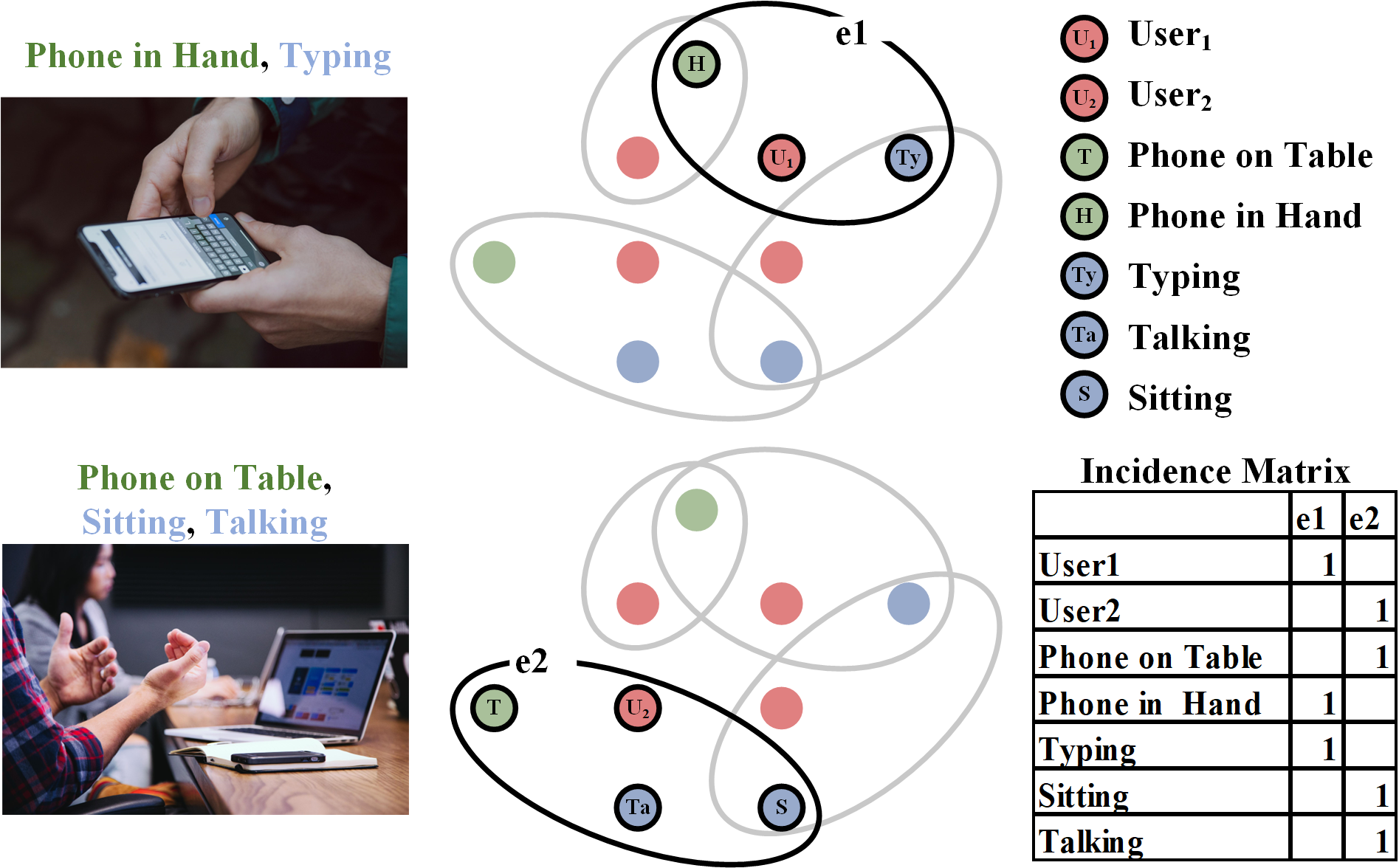}
    \caption{Real-world examples mapping into our heterogeneous hypergraph. A CA-HAR task has three types of nodes: users $u$, activities $a$, and sensor context $c$. We use red, green, and blue colors to represent these nodes with heterogeneity. The first example shows ${User}_{1}$ ($u_1$) is \textit{typing} ($a_{Ty}$) with \textit{phone in hand} ($C_H$). Thus, a hyperedge connecting three nodes $\{u_1, a_{Ty}, c_{H}\}$ is formed to represent the situation. Another example showcased scenarios where activities may co-occur: $User_2$ ($u_2$) is \textit{sitting} ($a_{S}$) and \textit{talking} ($a_{Ta}$) simultaneously with \textit{phone on table} ($c_T$). In this case, a hyperedge connecting four nodes $\{u_2, (a_{Ta}, a_{S}), c_{T}\}$ is needed to well-represent the situation. A corresponding incidence matrix is also shown to represent the subgraph with only these seven nodes and two hyperedges.}
    \label{fig:graph}
\end{figure}

Human Activity Recognition (HAR)~\cite{rault2017survey}, which involves recognizing human activities from sensor signals generated by Inertial Measurement Units (IMUs) attached to human bodies, is an important task in Context-Aware systems and has diverse real-world  applications~\cite{chen2018evaluating,lindqvist2011undistracted}. 
HAR has drawn great interest from both academia and industry domains due to the nearly ubiquitous ownership of smart devices that contain various types of sensors~\cite{vaizman2018context,ge2020cruft,ge2022qcruft,ge2023heterogeneous}. 
HAR is a challenging problem for several reasons. First, activities may co-occur (e.g., sitting and talking simultaneously, as shown in Fig.~\ref{fig:graph}), making HAR a challenging multi-label classification problem.
Secondly, some instances may have some missing labels (i.e., the ground truth of some labels is unknown).
Moreover, sensor signals for the same activity may vary significantly due to the impact of contextual factors such as phone placement~\cite{kang2022augmented,chang2020systematic} and user performance style~\cite{chen2020metier,bai2020adversarial}, as shown in Fig.~\ref{fig:contextNuser}. 

Inspired by these observations, recent research found it beneficial to incorporate the context of sensor signals and user information during data analyses and formulate the Context-Aware Human Activity Recognition (CA-HAR) task~\cite{zhang2022if,ouchi2013smartphone} to effectively boost HAR performance. In a CA-HAR task, a pattern recognition model infers both human activities being performed and the user's context from a set of sensor signals. In this paper, we focus on a neural network CA-HAR model to predict the user's current activity along with their context (i.e., smartphone placement), which are provided as labels in the dataset.

Prior CA-HAR work generally falls into three categories: 1) non-graph methods, 2) feature-dependent graph methods, and 3) feature-independent graph methods. Early prior work~\cite{alajaji2020deepcontext,vaizman2018context,ge2020cruft,ge2022qcruft} mostly utilized non-graph methods, where researchers performed handcrafted features and then directly applied existing machine learning methods for activity recognition without trying to construct a graph. To improve performance, other researchers proposed feature-dependent graph methods~\cite{martin2018graph,li2022disenhcn}, which designed specialized graph structures such as graphs derived from geographic features. A location-based graph improves HAR performance by factoring in the location where activities are performed. For instance, people are more likely to sleep at home or perform workouts in gyms. However, location-based graphs require that users agree to allow the collection of privacy-sensitive information such as GPS coordinates. Unfortunately, prior work has found that less than 50\% of users are willing to grant such access permission~\cite{dogrucu2020moodable}, limiting the applicability of location-based graph approaches. 

Feature-independent graph approaches are more general and do not rely on specific features within a dataset.  Instead,  graphs are formed solely based on three common elements across all datasets: 1) the user, 2) the context (placement of a device that eventually collects signals), and 3) the activity. Thus, label correlations can be learned across instances by assimilating message-passing algorithms with their graph neural network counterparts.

In this paper, we build upon this line of feature-independent approaches, and argue that the CA-HAR task is naturally expressed as a supervised graph learning task, where nodes are associated with the <user, context, activity+> tuples, and the sensor signals are graph edges. While inferencing, given the previously unseen test sensor signal as an edge in the graph, the CA-HAR task is to classify the edge (defined by its connecting nodes) through a similarity kernel~\cite{ge2023heterogeneous}. Such a CA-HAR graph differs from ordinary graph because it has both a hypergraph property (an edge can connect to more than two nodes) and a heterogeneity property (there are more than one type of nodes and edges), thus making it a special heterogeneous hypergraph. {Heterogeneous hypergraphs have been leveraged in other domains for real-world applications including location-based social network modeling~\cite{yang2019revisiting,yang2020lbsn2vec++}, document recommendation~\cite{zhu2016heterogeneous}, contagion analysis~\cite{st2022influential}, spam detection~\cite{sun2021heterogeneous}, and link prediction~\cite{fan2021heterogeneous}.} To the best of our knowledge, only a few works have explored creating feature-independent graphs for the CA-HAR task. {The most relevant prior research was conducted by Ge \etal~\cite{ge2023heterogeneous}, which proposed a hypergraph convolutional neural networks based approach and utilized separate linear projections to account for node-heterogeneity. } {We highlight two crucial ways in which their work differs from ours: }

\noindent {1) \textbf{The shallow-node-heterogeneity problem.} Using separate linear projections to map node representations into sub-spaces does not necessarily guarantee sufficient node-heterogeneity. The separate linear projections may still eventually lead to similar representations. A quantitive comparison in Sec.~\ref{sec:experiment} shows its sub-optimal performance, and a qualitative visualization in Fig.~\ref{fig:umapHHGNN} shows that it still has inter-mixing representations across different types of nodes.}

\noindent {2) \textbf{Missing labels that are common in in-the-wild CA-HAR data that result in missing hyperedges in our constructed graph, are not addressed.} We present evidence that this problem is not trivial by presenting hyperedge count statistics in Fig.~\ref{fig:edgecount} for two real-world datasets where over 9.5\% $\{u,c\}$ hyperedges in the \textit{WASH} dataset indicate a high activity missing rate, and over 42\% $\{u,a\}$ hyperedges in the \textit{Extrasensory} dataset indicates a high context missing rate. Motivated by the observation of missing labels, which commonly occur in CA-HAR data, resulting in missing hyperedges in our constructed graph, we designated this issue as an edge-heterogeneity problem that was accommodated by constructing corresponding sub-hypergraphs. }

{To address the problems mentioned above}, we propose a novel framework \textbf{D}eep \textbf{H}eterogeneous \textbf{C}ontrastive \textbf{H}yper-\textbf{G}raph \textbf{L}earning ({\model}) to address both node-heterogeneity and edge-heterogeneity in the heterogeneous hypergraph for the CA-HAR task with the following key pieces: 1) We address explicit node-heterogeneity with contrastive loss~\cite{hadsell2006dimensionality} where similar samples were pulled together, and dissimilar samples were pushed apart. In our case, we defined similar nodes as same-type nodes (i.e., both are users or activities, or contexts), while dissimilar nodes as nodes with heterogeneity (i.e., user-context pairs, user-activity pairs, or context-activity pairs). {The main reason for introducing contrastive loss is to ensure node-heterogeneity from a loss/objective regularization perspective, rather than only having shallow linear projections proposed by Ge \etal.}
2) We designed three specific hypergraph convolutional networks for the sub-hypergraphs denoted as $G_{(u,c,a)}$, $G_{(u,c)}$, and $G_{(u,a)}$, where each sub-hypergraph consists of hyperedges connecting to unique types of nodes, such that optimal layer parameters can be learned independently. {This is in contrast to the unified graph learning layer and one global hypergraph utilized by Ge \etal.} 
To the best of our knowledge, our work is the first to introduce contrastive loss and edge-heterogeneity toward solving the CA-HAR task. In extensive evaluation, {\model} consistently outperformed various state-of-the-art models on two real-life CA-HAR datasets. The further analysis illustrated how our contrastive loss better regularized the node embedding distributions, such that node-heterogeneity was well preserved and represented through Uniform Manifold Approximation and Projection for Dimension Reduction (UMAP) visualizations, thus enhancing model interpretability and explainability for human understanding.

\smallskip\noindent{This work makes the following key contributions:}
\squishlist

\item We cast the CA-HAR task as a supervised message-passing task trained using graph neural networks. 
We transform CA-HAR data to a heterogeneous hypergraph with three sub-hypergraphs accounting for different label-occurrence scenarios. Our CA-HAR graph explicitly encodes the relationships between node entities, as user performance style and context (e.g., phone placement) can enhance the performance of HAR~\cite{zhu2016heterogeneous,li2022disenhcn}.

\item We address node-heterogeneity in the CA-HAR graph by introducing a contrastive loss, which forces the same-type node embeddings to be close together while pushing nodes from different categories further away. 

\item We address edge-heterogeneity caused by missing labels in the CA-HAR graph by varying edge types based on their connecting nodes and leveraging different corresponding hypergraph convolution layers with separate parameters for capturing their unique message passing patterns.

\item We performed a rigorous evaluation of our novel {\model} based on the proposed node-heterogeneity and edge-heterogeneity methods and achieved superior performance against various baselines, including non-graph methods, ordinary graph methods, and prior shallow heterogeneity hypergraph methods, with a large margin where the average activity recognition improvements ranged from 5.8\% to 16.7\% on Matthews Correlation Coefficient (MCC) and 3.0\% to 8.4\% on Macro F1 scores. Further ablation study verified the non-trivial contribution of each novel component of our framework.
\squishend

\begin{figure}[htbp]
    \centering
    \small
    \subfloat[Work on an ordinary graph. E.g., GCN~\cite{kipf2016semi}.]{
        \includegraphics[width=0.27\linewidth]{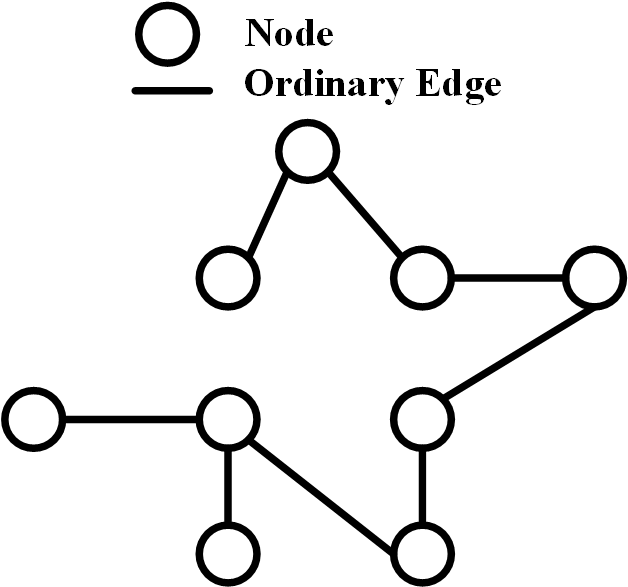}
        \label{fig:ordinarygraph}
    }
    \hspace{10pt}
    \subfloat[Work on hypergraph with node-heterogeneity. E.g., HHGNN~\cite{ge2023heterogeneous}.]{
        \includegraphics[width=0.29\linewidth]{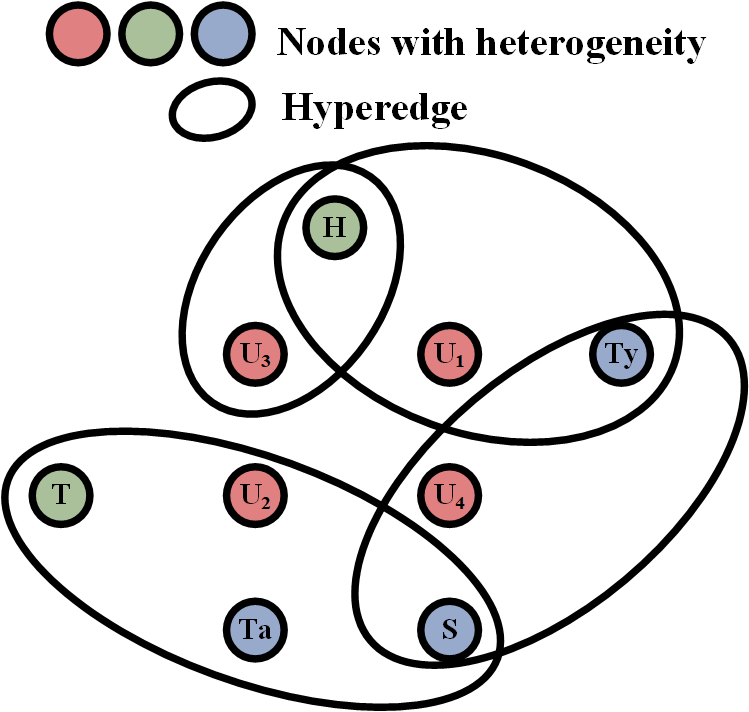}
        \label{fig:hypergraph}
    } \hspace{10pt}
    \subfloat[Work on an hypergraph with both node-heterogeneity and edge-heterogeneity (our approach).]{
        \includegraphics[width=0.3\linewidth]{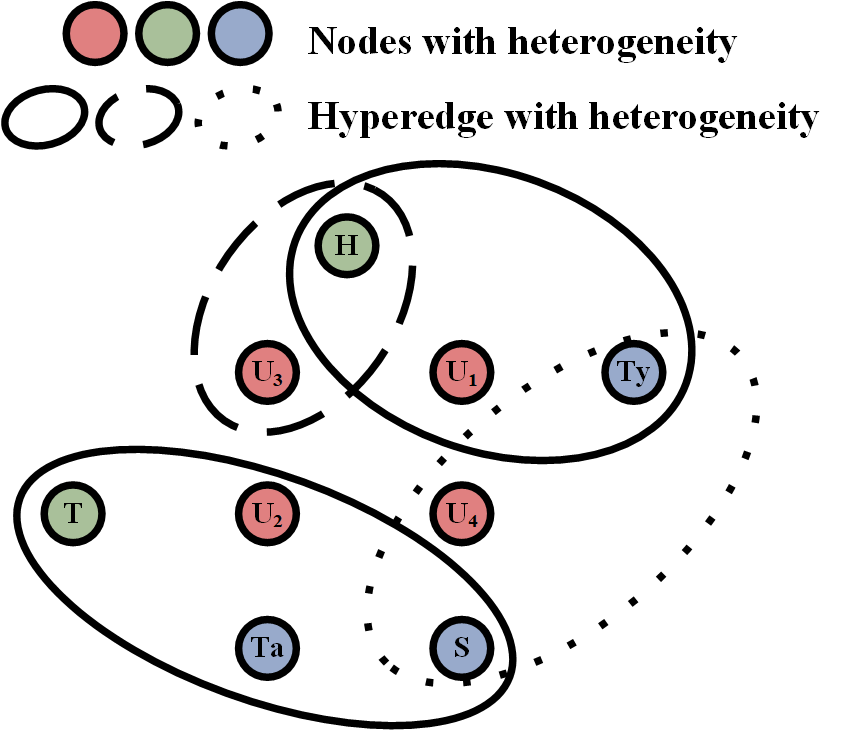}
        \label{fig:dhc-hgl}
    } \hspace{-1cm}
    \qquad
    \vspace{-3pt}
    \caption{{Prior work (Fig.~\ref{fig:ordinarygraph} and Fig.~\ref{fig:hypergraph}) constructs different CA-HAR graphs. Our approach (Fig.~\ref{fig:dhc-hgl}) considers both node-heterogeneity and edge-heterogeneity.  Intuitively, Fig.~\ref{fig:ordinarygraph} is a simplified version of Fig.~\ref{fig:hypergraph}, while Fig.~\ref{fig:dhc-hgl} further generalizes beyond Fig.~\ref{fig:ordinarygraph} and Fig.~\ref{fig:hypergraph}. Fig.~\ref{fig:illustration} presents an illustration with different sub-hypergraphs.}
    }
    \label{fig:compareHypergraph}
\end{figure}

\begin{figure}[htbp]
    \centering
    \vspace{-10pt}
    \subfloat[Sub-hypergraph $G_{(u,c,a)}$ contains hyperedges connecting all three types of nodes (user, phone placement and activity). E.g., $User_1$ is \textit{typing} with \textit{phone in hand}, and $User_2$ is \textit{sitting} and \textit{talking} with \textit{phone on table}.]{
        \includegraphics[width=0.30\linewidth]{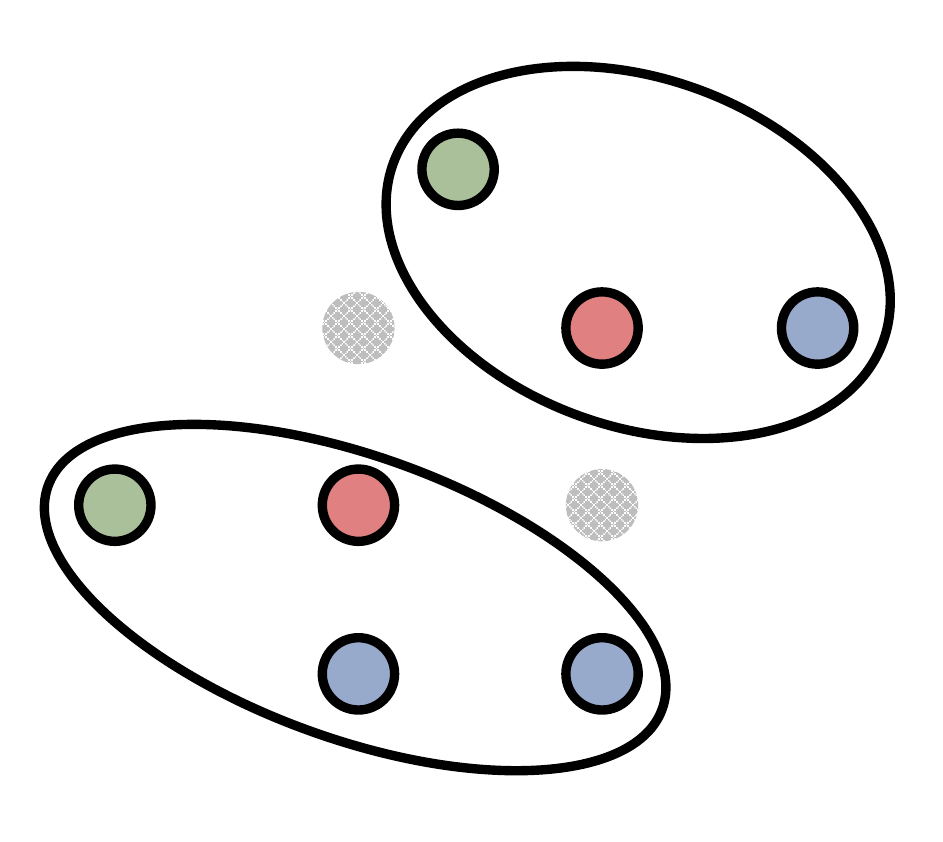}
        \label{fig:uppa}
    } \hspace{-0.5cm}
    \qquad
    \subfloat[Sub-hypergraph $G_{(u,c)}$ contains a hyperedge connecting user and phone placement nodes while activity nodes can be missing. E.g., $User_1$ holds \textit{phone in hand} and no activity label reported.]{
        \includegraphics[width=0.30\linewidth]{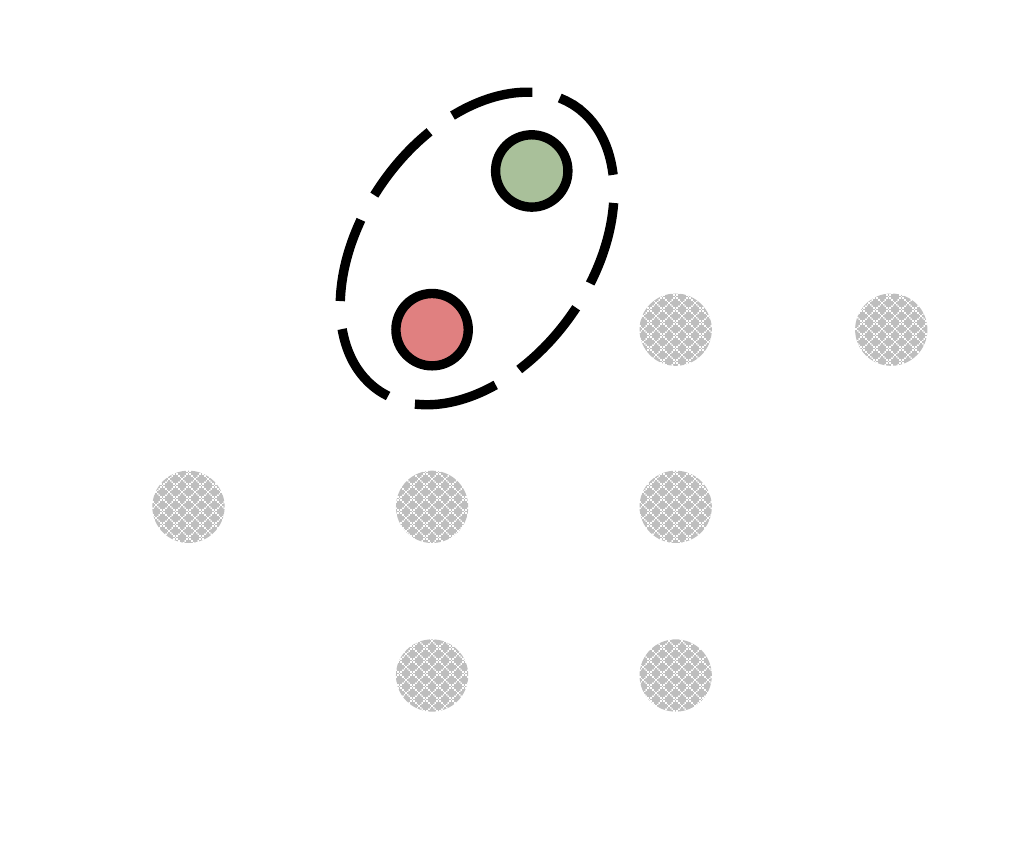}
        \label{fig:upp}
    }\hspace{-0.5cm}
    \qquad
     \subfloat[Sub-hypergraph $G_{(u,a)}$ contains a hyperedge connecting user and activity nodes while phone placement nodes can be missing. E.g., $User_3$ is \textit{typing} and \textit{walking} simultaneously, and no context label is reported.]{
        \includegraphics[width=0.30\linewidth]{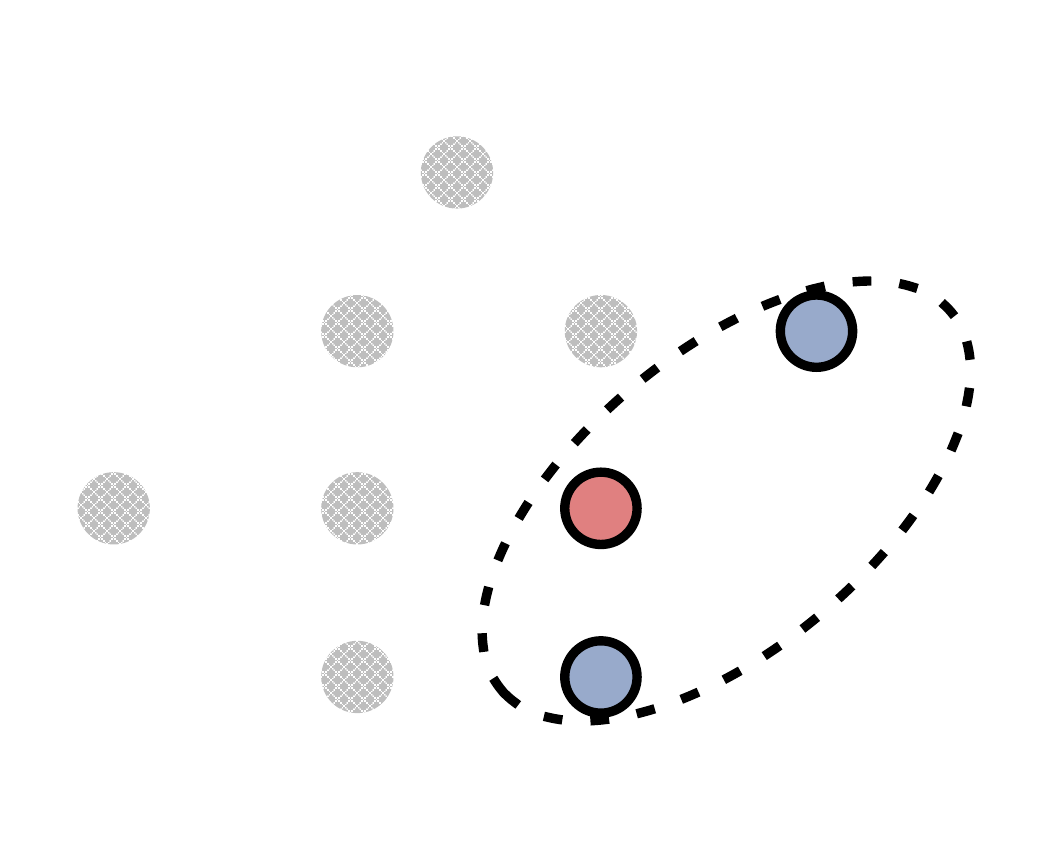}
        \label{fig:ua}
    }
    \vspace{-3pt}
    \caption{{The original heterogeneous hypergraph is further transformed into three sub-graphs that account for different types of hyperedges. In each sub-graph, node-heterogeneity is represented using different colors. However, only one type of hyperedge is incorporated. Thus, the sub-graphs capture graph information at a higher granularity.}
    }
    \label{fig:illustration}
\end{figure}
\section{Preliminary}
In this section, we first introduce the definitions around key concepts in graphs. We then show some illustrations around these definitions as well as real-world examples in the CA-HAR task.

\begin{definition}
\label{def:hhg}
(Heterogeneous Hypergraph~\cite{sun2021heterogeneous,ma2022learning}). A heterogeneous hypergraph $G = \{V, E, T_v, T_e, W\}$ contains vertices $V$ and hyperedges $E$, whereas $T_v, T_e$ represents the corresponding types of vertices and edges and $W$ denotes the edge weights. Below are the related concepts:
\squishlist
\item (Hyperedge in a hypergraph) Each hyperedge in a hypergraph can connect to more than two nodes, in contrast to ordinary graphs where each edge only connects to two nodes. Thus, hyperedges can be represented as set of connected nodes $e = \{v_k ~|~ v_{k} \in V\} \subseteq V$, for $e \in E$.
\item (Graph Heterogeneity) The hypergraph is heterogeneous when $|T_v| + |T_e| > 2$. \textit{Node-Heterogeneity} occurs when $T_v > 1$ and \textit{Edge-Heterogeneity} occurs when $T_e > 1$.
\item (Edge Weights) Typically, $W \in R^{|E|}$ is the matrix representing hyperedge weights.
\item (Incidence Matrix) The relationship between nodes and hyperedges can be represented by an incidence matrix $Inc \in R^{|V| \times |E|}$ with entries defined as:
\begin{equation}
    Inc(v,e) = 
\begin{cases}
    1, \text{if } v \in e \\
    0, \text{otherwise}
\end{cases}
\end{equation}
\squishend

\end{definition}

{We present real-world examples of the heterogeneous hypergraph properties of the CA-HAR task in Fig.~\ref{fig:graph}, a comparison of graphs formed in Fig.~\ref{fig:compareHypergraph}, and a high-level abstraction of three sub-hypergraphs in Fig.~\ref{fig:illustration}. 
Intuitively, Fig.~\ref{fig:ordinarygraph} is a simplified version of Fig.~\ref{fig:hypergraph}, while Fig.~\ref{fig:dhc-hgl} further generalizes beyond Fig.~\ref{fig:ordinarygraph} and Fig.~\ref{fig:hypergraph}. We found that real-world datasets contain edge-heterogeneity where edges may connect to different types of nodes, namely $G_{(u,c,a)}$ in Fig.~\ref{fig:uppa}, $G_{(u,c)}$ in Fig.~\ref{fig:upp}, and $G_{(u,a)}$ in Fig.~\ref{fig:ua}. {More specifically, 1) each hyperedge in $G_{(u,c,a)}$ connects to all three types of nodes simultaneously (i.e., user, context, and activity nodes); 2) the hyperedges in $G_{(u,c)}$ only connect to user and context nodes, and 3) $G_{(u,a)}$ contains hyperedges that are associated with only user and activity nodes. }Directly applying pattern recognition models on Fig.~\ref{fig:hypergraph} without factoring in the existence of edge-heterogeneity in Fig.~\ref{fig:uppa}, \ref{fig:upp}, and \ref{fig:ua} can yield sub-optimal performance, as the message-passing patterns through these different hyperedges may vary. The key reason for addressing edge-heterogeneity is that it accounts for missing labels common in in-the-wild CA-HAR datasets by modeling hyperedges at a higher granularity. The inferior performance of the variant of our proposed model and HHGNN~\cite{ge2023heterogeneous}, which both applied a unified Graph Neural Network(GNN) without edge-heterogeneity, presented in Sections~\ref{sec:results} and~\ref{sec:analysis}, support this claim.}

\section {Related Work}
\label{sec:relatedwork}
Context-aware human activity recognition (CA-HAR) is an emerging task in academia and industry~\cite{vaizman2018context,ge2020cruft,ge2022qcruft,ge2023heterogeneous}, as CA-HAR is crucial in many real-world applications, such as healthcare~\cite{ouchi2013smartphone}, smart homes~\cite{bianchi2019iot}, and biometric authentication~\cite{zou2020deep}. Prior works focused on recognizing human activities given various types of data, such as images~\cite{shi2020skeleton} and videos~\cite{liu2023amir}, wearable sensors~\cite{kang2022augmented}, smartwatches~\cite{vaizman2018context}, and smartphone sensors~\cite{vaizman2018context,ge2020cruft,ge2022qcruft,ge2023heterogeneous}. This work focuses on leveraging smartphone sensor signals for recognizing human activities, but one may extend our work to other data sources or modalities with minimum customization efforts. 

More recently, many deep learning models were proposed for Context-Aware Human Activity Recognition (CA-HAR), and they mainly fall into the following categories:

\smallskip\noindent\textbf{Non-graph CA-HAR methods.}
Existing non-graph HAR methods mainly focused on feature engineering, utilizing predictive handcrafted features extracted from raw signals~\cite{vaizman2018context,ge2023heterogeneous}, feature mining (i.e., directly learning from raw signals)~\cite{yao2018rdeepsense}, or combining both types of features~\cite{alajaji2020deepcontext,ge2020cruft,ge2022qcruft}. In terms of deep learning frameworks, the most common methods are Multi-Layer perceptron (MLP) on handcrafted features~\cite{vaizman2018context}, Convolutional Neural Networks (CNN) that capture spatial correlations~\cite{ge2020cruft,ge2022qcruft,bai2020adversarial}, Recurrent Neural Networks (RNNs, LSTMs) on learning temporal dependencies~\cite{ge2020cruft,ge2022qcruft}, and uncertainty measurements for mitigating data noise problems and missing label problems~\cite{yao2018rdeepsense,ge2020cruft,ge2022qcruft}. Despite exploring multiple perspectives in the HAR task, the common problem within non-graph HAR methods is they do not explicitly model the inter-entity (users, context, and activities) dependencies.

\smallskip\noindent\textbf{Feature/Sensor dependent graph HAR methods.}
Graph Neural Networks (GNNs) have been widely applied in HAR, particularly on videos in the Computer Vision domain, including learning Actor Relation Graph connections for group activity recognition~\cite{wu2019learning} and utilizing spatial-temporal graphs for skeleton-based action recognition~\cite{shi2020skeleton}. Another sensor-based HAR method proposed by Martin et al.~\cite{martin2018graph} built two personalized mobility graphs (transition frequency and Euclidean distances as edge weights) using GPS sensors and applied two GCNs to recognize human activities. The derivation of the graphs depends on the availability of specific features/sensors, e.g., GPS sensors, which might violate user privacy and require users' permission. Unfortunately, less than 50\% of users were willing to grant access to GPS sensors~\cite{dogrucu2020moodable}. In summary, the feature/sensor-dependent graph method is a double-edged sword: they enabled models to better learn spatial and temporal correlations through a graph perspective; however, such a method also relies highly on the specific feature's existence, which limited their applicable scope in the HAR domain.

\smallskip\noindent\textbf{Feature independent graph methods.}
Unlike feature-dependent graph methods, feature-independent graph methods can construct graphs without requiring external information. Some prior work utilized fully connected graphs~\cite{mohamed2022har,zheng2022meta} for activities where edges are weighted through similarity kernels. Others formed graphs based on K-nearest-neighbours~\cite{peterson2009k}, i.e., they connect each node with the TopK nearest neighbors based on the feature embeddings~\cite{liu2021human}. Despite different designs for their downstream neural networks, they share drawbacks wherein their ordinary graphs do not contain hyperedges. Thus, they do not factor in the co-occurrence of more than two entities. Moreover, although node-heterogeneity can be incorporated into their frameworks, edge-heterogeneity was mostly overlooked.

{The most relevant work to our research is from Ge~\etal~\cite{ge2023heterogeneous}. The authors transformed context-aware human activity visit patterns into a corresponding heterogeneous hypergraph. 

To address the node-heterogeneity characteristic, the user, phone placement, and activity nodes were treated as different types of nodes with separate fully connected layers in-between multiple HyperGraph Convolution Layers (HGC)~\cite{bai2021hypergraph}. Despite their encouraging work, several problems remain unresolved: 1) \textit{Shallow-node-heterogeneity}: Having separate linear projections for different types of nodes essentially sends their representation into sub-spaces, but there is no guarantee on the distribution of the representations. In fact, in the worst case scenario, separate linear projections can learn similar weights, thus behaving in the same way as a single unified fully connected layer without node-heterogeneity. 2) \textit{Overlooking edge-heterogeneity}. As a direct result, the missing label problem that is common in in-the-wild CA-HAR datasets is not properly handled.}

\smallskip\noindent\textbf{{Prior Graph Neural Network (GNN) research: }}

{Graph Neural Networks (GNNs) have attracted a lot of attention. The key idea of GNNs is to define graph nodes and egdes, that are used to generate node representations by aggregating information from the node's neighbors. For example, GCN~\cite{kipf2016semi} learns node embeddings from its first-order neighbors using the Kipf normalized aggregation. GraphSAGE~\cite{hamilton2017inductive} scales to large datasets by utilizing neighborhood sampling and neural networks such as Long Short Term Memory (LSTM) models to aggregate neighbor's information. GAT~\cite{velivckovic2017graph} introduces attention mechanisms that specify different importance weights to different neighbors.}
{Recently, more GNNs have proposed various mechanisms that adapt GNNs to handle real-world applications better, including higher-order relationships between objects, and different types of nodes and edges. For instance, hyperConv~\cite{bai2021hypergraph} incorporates higher-order relationships by allowing information propagation between multiple nodes concurrently. HetGNN~\cite{zhang2019heterogeneous} was proposed for various graph mining tasks, including link prediction, recommendation, node classification, and clustering. It addressed node-heterogeneity by grouping different types of neighbors sampled using a Random Walk with Restart (RWR) algorithm and designing a specific module to extract heterogeneous content from each neighboring node. However, using the same restart probability for all nodes limits the expressiveness of the sampled neighbors, further limiting the expressiveness of the aggregated node representation.
HERec~\cite{shi2018heterogeneous}, originally proposed for recommender systems, addresses data heterogeneity by using a meta-path based random walk strategy to convert a heterogeneous graph into a homogeneous graph. However, manually defining meta-paths to model these semantic relationships in heterogeneous graphs heavily relies on the quality of the designer's domain knowledge.} 

\smallskip\noindent\textbf{Novelty of our work in relation to prior work: } 
Our work differs from prior works (especially the HHGNN model proposed by Ge~\etal~\cite{ge2023heterogeneous}) in at least two crucial ways: 
1) we address node-heterogeneity with a contrastive loss objective, which pulls nodes from the same type closer by minimizing their distance and nodes from different types far apart by maximizing their distance. {The key reason for using a contrastive loss function is that it explicitly regularizes the node representations, rather than implicitly projecting node representations into sub-spaces. Consequently, we are able to guarantee node-heterogeneity in a distance-based manner. Specifically, nodes of the same type have similar representations, while nodes of different types are distant from each other.}
It not only improved the model performance but also enhanced human-level understanding of the learned node representation distributions.
and 2) we explicitly address edge-heterogeneity in our hypergraph framework design by distinguishing hyperedges via its connected nodes and designing separate hypergraph convolution layers to allow independent message propagation. {The key rationale for the design is: by addressing edge-heterogeneity, we provide a solution that accounts for missing labels which are common in in-the-wild CA-HAR datasets.}

\begin{figure*}[t]
\centering

    \includegraphics[width=1.\linewidth]{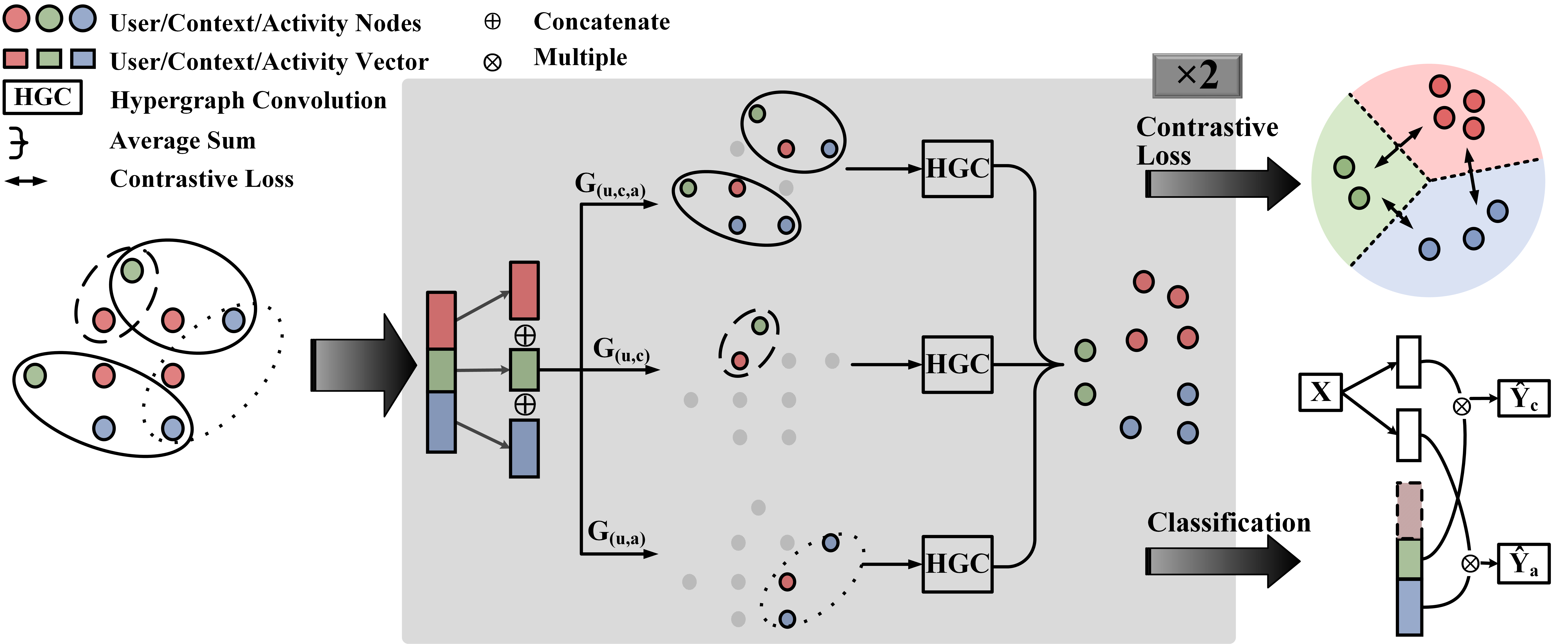}

    \caption{Overview of our {\model} Framework. {It consists of two key components: the graph learning module for label encoding and the classification module for signal encoding. Given the heterogeneous hypergraph formulated (each node is defined as the user, context, and activity tuple), the model updates node representations using a GNN while factoring in node-heterogeneity (via separate projections and custom HGC layers), and edge-heterogeneity (via split subgraphs). During classification, the learned label encoding is utilized to infer connected nodes for the given signal. The objective loss function combines BCE loss for multi-label classification and contrastive loss that handles node-heterogeneity.
    }}
    \label{fig:framework}

\end{figure*}

\section{Proposed Framework}
\label{sec:framework}
We formulate the CA-HAR problem in Section~\ref{sec:problem}. Each component of our framework is then described in Section~\ref{sec:model}. A conceptual visualization of our proposed {\model} is shown in Fig.~\ref{fig:framework}. 

\subsection{Problem Formation}
\label{sec:problem}
CA-HAR is a supervised multi-label classification task where a dataset $D$ contains data instances $x \in X$, the corresponding data labels $y \in Y$, and the users involved $u \in U$:
\begin{equation}
    D = \{(x,y,u) ~|~ x \in R^{M}, y \in [0,1]^H, u \in [1,2,..., |U|]\}
\end{equation}
where $M$ is the dimension of the sensor signal, and $H$ is the number of labels per instance, which is equivalent to the total number of contexts and activities.
\begin{equation}
    H = {|context|} + {|activity|}
\end{equation}
A CA-HAR model $m$ learns an effective mapping between data instance X and data label Y:

\begin{equation}
    m: X \rightarrow Y, m(x_i) = y_i, 0 \le i < |X|
\end{equation}
Under a heterogeneous hypergraph transformation, each sensor signal $x \in X$ can be associated with three types of elements: users, context, and activities, where the last two belong to Y. If each sensor signal is considered as a hyperedge and the three types of elements are considered as graph nodes, the original dataset D will be transformed into a graph G. 
\begin{equation}
\begin{aligned}
V &=\{U\}+\{Y\} \\
E &= \{e_i\}_{0 \le i < |E|}\\
    G &= (E,V)
\end{aligned}
\end{equation}
Each edge $e$ can also be uniquely represented by their connecting nodes as a set:
\begin{equation}
    e_i = \{v_{ik} ~|~ v_{ik} \in V, k=0,1,...\}
\end{equation}
Now that the new model $m_G$ is given a hyperedge $e$ as input, it tries to infer the set of nodes connecting to hyperedge $v$:
\begin{equation}
    m_G(e) = v, e \in E, v \subseteq V
\end{equation}

\subsection{Network Design}
\label{sec:model}
{Our network is composed of two key components: the graph learning module for label ($Y$) encoding, and the classification module for signal ($X$) encoding. At a high level, the graph learning module encodes the labels based on their co-occurrence relationship in a message-passing fashion, while the classification module encodes the data representations. {During training, node representations are updated by comparing the label encodings with all training data/signal encodings. During inference, the final prediction is made via a dot product between label encodings and the test data/signal encodings, thus deriving the labels that are most similar to a given signal.}}

{To the best of our knowledge, one of the many advantages of introducing graph learning into the CA-HAR domain is that we are able to transform simplified binary label encoding into more expressive vector representations. This goal is achieved with the help of aggregated signal information. Moreover, we are able to enhance the signal specificity of data encoding. Thus, both the macro-level and micro-level information from a given dataset can be obtained.}

{Our main contributions, however, are based on two key insights in improving the graph learning expressiveness that were overlooked in prior work. First, we address edge-heterogeneity to resolve the problem of missing labels as described in earlier sections.  Secondly, we ensure node-heterogeneity using explicit, objective contrastive loss regularization, rather than simpler separate linear projections that may degrade to projected nodes that do not capture heterogeneity. We describe the details of our network design in the following subsections, and demonstrate their outstanding performance in Section~\ref{sec:experiment}.}

\subsubsection{Graph Learning}
\label{sec:gl}
Given the formulated graph $G=(E,V)$, each node is initialized to the average of all instances connected to it:
\begin{equation}
    Emb_{v} = mean(x_{t}), \text{where } y_{t} == 1, v \in V
    \label{eq:nodeInit}
\end{equation}
The node embedding is arranged in the order of users (u), context (c), and activities (a)
\begin{equation}
    V_g = \oplus [V_u, V_c, V_a]
    \label{eq:split}
\end{equation}
{where $\oplus$ represents the concatenation operation.}
The hyperedges are initialized to the average of sensor signal instances that have the same connecting nodes: 
\begin{equation}
\begin{aligned}
e &= \{v_{k}\}_{k = 0,1,...}, & \text{where } e \in E, v_{k} \in V, \{v_{k}\} \subseteq V \\
Emb_{e} &= mean(x_{kj}), & \text{for all } \cup \{u_{kj}, y_{kj}\}_{j=0,1,...} == \{v_{k}\}
\end{aligned}
\end{equation}
Given the graph G and initialized vector representations as inputs, we try to learn node representations using the graph network. To factor in node-heterogeneity, separate linear projections are then applied, followed by non-linear activations and dropout on each type of node.
\begin{equation}
\begin{aligned}
    V^{'}_g &= \oplus [V^{'}_u, V^{'}_c, V^{'}_a] \\
    V^{'}_t &= \delta \cdot \alpha \cdot l_t(V_t), t \in (u,c,a) \\
\end{aligned}
\label{eq:nodeHetero}
\end{equation}
where $\delta$ represents a dropout operation, $\alpha$ is a non-linear activation function, and $l_u, l_c, l_a$ are three linear projections corresponding to different node types.
The hyperedges are then split into three groups based on the different types of nodes they are connected to, namely:
\squishlist
\item $\{u,c,a\}$ edges: edges connect all three types of nodes
\item $\{u,c\}$ edges: edges connect only user and context nodes, and no activity was provided
\item $\{u,a\}$ edges: edges connect only user nodes and activity nodes, and no context was provided
\squishend
The distribution of each type of hyperedge is shown in Fig.~\ref{fig:edgecount} for two real world datasets (i.e., \textit{WASH Unscripted} and \textit{Extrasensory}, with detailed descriptions in Section~\ref{sec:dataset}). 
Thus, there are three subgraphs $G_{(u,c,a)}$, $G_{(u,c)}$,  and $G_{(u,a)}$. {Each component goes through a different hyperConv layer~\cite{bai2021hypergraph} with non-linear activation and dropouts. Their results are aggregated into the whole graph through a summation function to form final node representations.}
\begin{equation}
\begin{aligned}
    & V^{''}_{gs} = \delta \cdot \alpha \cdot {HGC}_{s}(V^{'}_{gs})\\
    & V^{''}_g = aggr(V^{''}_{gs}) \\
    & s \in \{(u,a), (u,c,a), (u,c)\} 
    \end{aligned}
    \label{eq:hyperConv}
\end{equation}
The combination of Eq.~\ref{eq:split}, \ref{eq:nodeHetero} and~\ref{eq:hyperConv} form one full layer of our heterogeneous hypergraph convolutional layer. In practice, several such layers can be stacked together for learning multi-hop neighborhood graph properties
such that long-term dependencies may be better captured. Take the user node $U_1$ in Fig.~\ref{fig:hypergraph} as an example: one heterogenous hypergraph convolution layer can aggregate information from its 1-hop neighbors (i.e., node \textit{H} and \textit{Ty}) while with two layers, we can aggregate additional information from node $U_3$, $U_4$ and \textit{S}, where user node $U_3$ can be considered as a similar user of $U_1$, given the common context they share (i.e., node \textit{H}). Thus the learned node embedding can capture richer information about the graph structure, which can further benefit the HAR task.

\subsubsection{Classification}
\label{sec:nodeClassification}
After learning effective node representations $V^{''}_g$, the model can infer connected nodes given hyperedge representations $x$ (i.e., sensor feature). This procedure can be considered an edge-type classification problem. First, both node representations and the sensor signal edge are projected into the same dimensions. Then the dot product
is computed for comparing vector similarities between a given sensor signal and nodes. As activities may co-occur, a sigmoid transformation on each dot product result is adopted, rather than the softmax function across results. The sigmoid results are treated as probabilities of each label as connecting nodes.
\begin{equation}
    \begin{aligned}
    x^{'}_t &= \delta \cdot \alpha \cdot l_{xt}(x) \\
    V^{''}_{gt} &= \delta \cdot \alpha \cdot l_{vt}(V^{''}_g) \\
    \hat{Y} &= \oplus [\sigma \cdot V^{''}_{gt} \otimes X^{'}_t], t \in \{c,a\} \\
    \end{aligned}
\end{equation}
where $\otimes$ represents the dot product operation and $\sigma$ represents the sigmoid function. {$l_{xt}$ and $l_{vt}$ are two linear projections to map node and edge representations into the same dimension, such that a dot product can be applied to the two components.}

\subsection{Objective Loss Function}
\label{sec:object}
We now design a loss function that has two components: 1) A muti-label classification loss with  label-wise Binary Cross Entropy (BCE) loss $L_{bce}$ and 2) A node-heterogeneity contrastive loss $L_c$:
\begin{equation}
\begin{aligned}
    & \underset{\theta_g, \theta_c}{\min} &L &= \underset{\theta_g, \theta_c}{\min} ~L_{BCE}(\hat{Y}, Y) +  \underset{\theta_g}{\min} ~L_{c}(V^{''}_g) \\
    & \text{where} & V^{''}_g &= \theta_g(V_g), \hat{Y} = \theta_c(V^{''}_g, x)
\end{aligned}
\end{equation}
where BCE loss is commonly adopted as:
\begin{equation}
L_{BCE} = \frac{1}{N}\sum_{n=1}^N\sum_{c=1}^C[-\omega_{n,c}(y_{n,c}log\hat{y}_{n,c} +(1-y_{n,c})log(1-\hat{y}_{n,c}))]
    \label{eq:bceloss}
\end{equation}
$N$ is the number of instances within each batch and $C$ is the number of classification labels.

\begin{figure*}[t]
\centering

    \includegraphics[width=.9\linewidth]{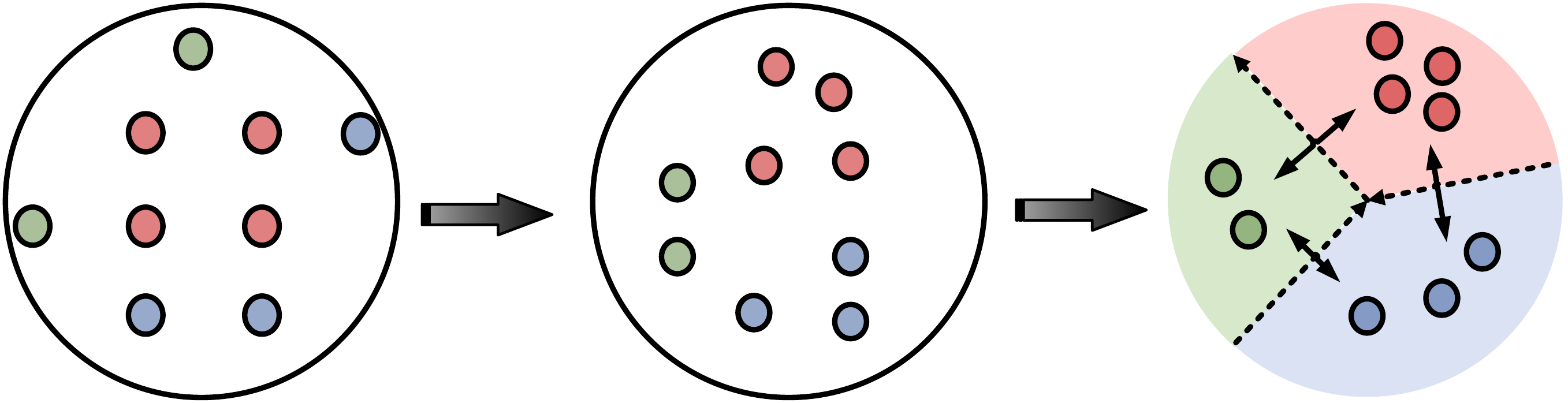}

    \caption{Illustration of how contrastive loss improves learning of node embeddings. Left: initial node embedding formed as Eq.~\ref{eq:nodeInit}. Mid: learned node embedding w/o contrastive regularization. Right: learned node embedding w/ contrastive regularization.}
    \label{fig:contrastiveloss}

\end{figure*}

To explicitly represent node-heterogeneity, we introduced a contrastive loss on node embeddings (Eq.~\ref{eq:contrastloss}), which essentially pulls nodes of the same type closer and pushes nodes of different types further away (Fig.~\ref{fig:contrastiveloss}). It calculates node embedding similarity pairwisely and pulls nodes of the same type closer by maximizing the similarity between them and vice versa. As there are a limited number of nodes (|U|+|C|+|A|), we adopted fully contrastive loss~\cite{hadsell2006dimensionality} that iterates over all node pairs rather than sampling-based loss~\cite{khosla2020supervised,ramamurthy2022cogax} which performs calculations based on sampled positive and negative pairs: 

\begin{equation}
    L_{c} = \frac{1}{H(H-1)}\sum_{V_i, V_j \in V^{''}_g, i \neq j} W(i,j) \cdot f(V_i, V_j)
    \label{eq:contrastloss}
\end{equation}
where $f$ is the distance function based on cosine similarity, and $W$ is a weight function:
\begin{equation}
    W(i,j) =
    \begin{cases}
        \lambda_1, & \text{if } I(i,j) == 1 \\
        -\lambda_2, & \text{otherwise}
    \end{cases}
\end{equation}
$I$ is the node type identification function. It outputs 1 if two nodes have the same type. Otherwise, it outputs -1. $\lambda_1, \lambda_2$ are hyperparameters whose optimal values can be found using grid search.

\begin{table}[htbp]
    \caption{Context-aware Human Activity Datasets Information. The Accelerometer, Gyroscope, and Magnetometer are 3-axis sensors. If the sample rate is unspecified, the sensor is sampled once per example.}
	\centering
    \small
\scalebox{.9}{
	\begin{tabular}{c||c|c |c | c | c }\hline
       \toprule
        \textbf{Dataset}    &\textbf{\#Instances}    &   \textbf{\#Participants}  &\textbf{\#Features} &\textbf{Common Sensors}    &\textbf{Unique Sensors}\\
        \midrule
        \textbf{WASH}&\num[group-separator={,}]{7773479} &  \num[group-separator={,}]{108}&  \num[group-separator={,}]{139}  & \multirow{3}{*}{\shortstack{Accelerometer(40Hz), Gyroscope(40Hz)\\Location, Magnet(40Hz)\\Env. Measure, Phone State}} & Response\\
        \cmidrule{1-4}
        \cmidrule{6-6}
        \multirow{2}{*}{\textbf{Extrasensory}}   & \multirow{2}{*}{\num[group-separator={,}]{6355350}} &  \multirow{2}{*}{\num[group-separator={,}]{60}}&  \multirow{2}{*}{\num[group-separator={,}]{170}}   &    &  Gravity\\
        &&&&&Audio(46Hz)\\
        \bottomrule
	\end{tabular}}
	\label{tab:statistic}
\end{table}

\section{Experiments}
\label{sec:experiment}
\subsection{Context-aware Human Activity Recognition Datasets}
\label{sec:dataset}
{We evaluate {\model} on two unscripted context-aware human activity recognition datasets, namely \textit{WASH}\footnote{\url{https://tinyurl.com/8wvrhr7k}} and \textit{Extrasensory}~\cite{vaizman2017recognizing}}. Unscripted datasets were collected in-the-wild, with participants running a data-gathering app on their smartphones as they lived their lives and provided context-aware activity labels periodically. 
Unscripted datasets are realistic and can provide better insights into participants' real-world behavior patterns. However, such datasets are quite noisy with user-provided labels that may be missing, wrong or conflicting with each other. Additionally, on different phone types, not all sensors are always available. Sensor readings may also be missing for several reasons, including weak signals and the fact that participants sometimes turn off their phones to save power or not give permission to collect data from certain sensors (e.g., GPS for privacy reasons). We describe pre-processing methods on the two datasets in Section~\ref{sec:preprocessing}.
Detailed statistics of these datasets are presented in Table~\ref{tab:statistic}. The \textit{Extrasensory}~\cite{vaizman2017recognizing} dataset collected 20 seconds of sensor measurement per minute from smartphones and smartwatches. 60 participants from diverse ethnibackgrounds, including 34 females and 26 males, were recruited in the study for approximately one week. 
\textit{WASH} followed a similar data collection and labeling methodology as \textit{Extrasensory}, 108 users participated in a data collection study for about two weeks. Context and activity labels collected by both datasets' are listed in Table~\ref{tab:labels}. We selected 17/51 labels from \textit{Extrasensory} that are similar to labels in the \textit{WASH} dataset to ensure a fair comparison.

\begin{table}[t]
    \caption{Context-aware Human Activity Labels.}

	\centering
    \small
\scalebox{.9}{
	\begin{tabular}{c||c|c |c | c |c| c}\hline
       \toprule
        \textbf{Dataset}    &\textbf{\shortstack{Common\\Context Label}}    &   \textbf{\shortstack{Unique\\Context Label}}  & \textbf{\shortstack{\#Context\\Label}}&\textbf{\shortstack{Common\\Activity Label}} &\textbf{\shortstack{Unique\\Activity Label}}   & \textbf{\shortstack{\#Activity\\Label}}\\
        \midrule
        \multirow{3}{*}{\textbf{WASH}}  &\multirow{6}{*}{\shortstack{In Pocket\\In Hand\\ In Bag}}&\multirow{3}{*}{\shortstack{On Table-Face Down\\ On Table-Face Up}}  &\multirow{3}{*}{5}& \multirow{6}{*}{\shortstack{Lying Down, Sitting\\ Walking, Sleeping\\ Standing, Running\\ Stairs-GoingDown\\ Stairs-Going Up\\ Exercising}}&\multirow{3}{*}{\shortstack{Talking On Phone\\ Bathroom\\ Jogging, Typing}} &\multirow{3}{*}{12}\\
        &&&&&&\\
        &&&&&&\\
        \cmidrule{1-1}
        \cmidrule{3-4}
        \cmidrule{6-7}
        \multirow{3}{*}{\textbf{Extrasensory}}& & \multirow{3}{*}{On Table} &\multirow{3}{*}{4}  &  &\multirow{3}{*}{\shortstack{Talking\\ Bath-Shower\\ Toilet}} & \multirow{3}{*}{13}\\\
        &&&&&&\\
        &&&&&&\\
        \bottomrule
	\end{tabular}}

	\label{tab:labels}
\end{table}

\subsubsection{Dataset Pre-processing}
\label{sec:preprocessing}
First, we resolved label conflicts by removing problematic data instances falling into the following cases: 1) Users provided more than one phone placement label resulting in co-occurring phone placement. Phone placements are mutually exclusive in the real world as a phone can be carried in only one position (e.g. in hand, back or coat pocket) at a time. 2) Multiple conflicting activity labels that cannot be performed simultaneously were provided by the study participant (co-occurred). For example, a smartphone user cannot sleep and run at the same time. 

Next, the raw multi-sensor signals were segmented into equal-sized time windows with a duration of 3 secs and a 1.5 sec step size, yielding \num[group-separator={,}]{7773479} and \num[group-separator={,}]{6355350} instances (Tab.~\ref{tab:statistic}), respectively. Next, the features extracted in many prior similar works were extracted~\cite{vaizman2018context,ge2022qcruft,ge2020cruft,ge2023heterogeneous} subject to sensor availability in both datasets. Finally, we generated 139 and 170 handcrafted features after removing identical features, respectively. 

Lastly, we normalized features in the train set into a range of 0 to 1 using Eq.~\ref{eq:norm} and then applied it to the test and validation sets.
\begin{equation}
    z = (x-\mu)/s
\label{eq:norm}
\end{equation}
where $\mu$ and $s$ are the mean and standard deviation of features in the training set. $x$ and $z$ are the original and transformed features, respectively. 
\begin{figure}[t]
\centering

    \includegraphics[width=.55\linewidth]{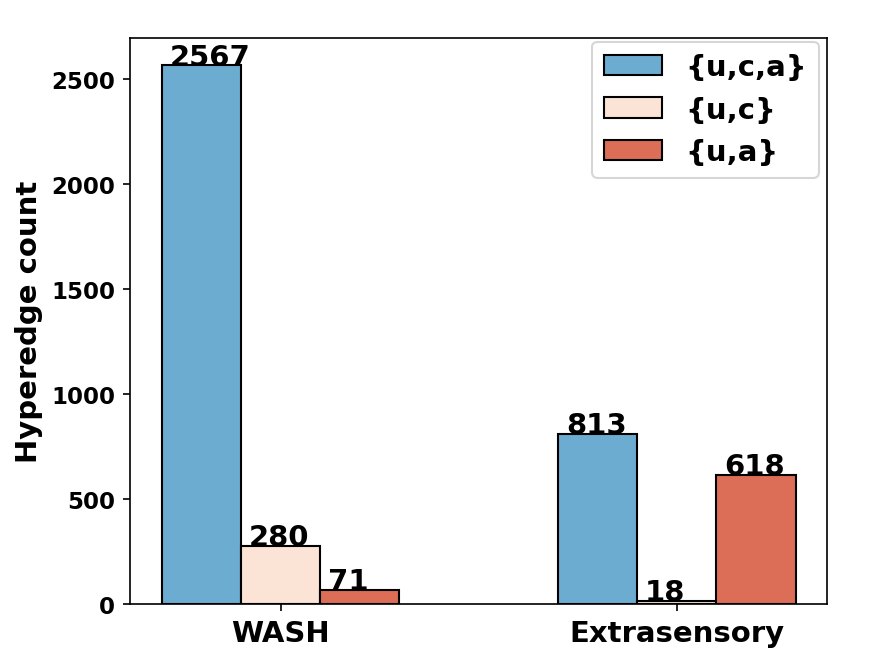}

    \caption{Hyperedge distribution on two datasets. {Specifically, $\{u,c,a\}$ represents hyperedges simultaneously connecting user, context and activity nodes, $\{u,c\}$ represents hyperedges connecting only user and context node, and $\{u,a\}$ represents hyperedges connecting only user and activity node.}}
    \label{fig:edgecount}

\end{figure}
As previously mentioned, the normalized features were used to initialize our graph with the aggregated mean function. The missing values in the initialized graph were filled with zeros. The number of unique hyperedges is reported in Fig.~\ref{fig:edgecount}. It is instructive to note that in prior works, hyperedges were treated as being similar (no heterogeneity). In contrast, edge-heterogeneity was incorporated into our framework.

\subsection{Baseline HAR Models}

\begin{table}[tbp]
    \caption{{Comparison of existing CA-HAR methods.}}
	\centering
    \small
\scalebox{.9}{
	\begin{tabular}{c|r||c|c | c | c | c}\hline
       \toprule
        \multirow{2}{*}{\textbf{Category}} &\multirow{2}{*}{\textbf{Model}}     & \multirow{2}{*}{\shortstack[c]{\textbf{Handcraft}\\\textbf{Feature}}} &   \multirow{2}{*}{\textbf{Methods}}  &\multirow{2}{*}{\textbf{Hypergraph}} & \multicolumn{2}{c}{\textbf{Heterogeneity}} \\
        & & & & & \textbf{Node} & \textbf{Edge} \\
        \midrule
        \multirow{3}{*}{Non Graph-based method} &
        \textbf{CRUFT}      &  \checkmark   & Deep Learning         &   -           &   -   &   -   \\
        
        &\textbf{LightGBM}   &   \checkmark  & Machine Learning      &   -           &   -   &   -   \\
        &\textbf{ExtraMLP}   &   \checkmark  & Deep Learning         &   -           &   -   &   -   \\
        \midrule
        \multirow{3}{*}{Graph-based method}
        &\textbf{GCN}        &   \checkmark  & Deep Learning    &               &       &       \\
        &\textbf{HHGNN}      &   \checkmark  & Deep Learning    & \checkmark    & \checkmark    &   \\
        &\textbf{\model}     &   \checkmark  & Deep Learning    & \checkmark    & \checkmark    & \checkmark    \\
        \bottomrule
	\end{tabular}}

	\label{tab:compare}
\end{table}

As part of our rigorous evaluation, we compared our proposed {\model} performance to several state-of-the-art baselines, including non-graph based models, ordinary graph-based models, and heterogeneous hypergraph models. {Table~\ref{tab:compare} summarizes the key differences between them.} We provide brief descriptions of models with rationale on why they were selected in the following content.
\squishlist
\item \textbf{CRUFT~\cite{ge2020cruft}:} is a state-of-the-art non-graph-based framework that achieved some of the best results till date. It jointly learns using two branches: one MLP branch analyzing handcrafted features and one CNN-BiLSTM branch analyzing raw accelerometer and gyroscope data. It exploited temporal correlation among instances by learning multiple consecutive samples together. CRUFT also incorporated an uncertainty estimation module to deal with noisy CA-HAR data collected in-the-wild. The random splitting of our data may undermine CRUFT performance as temporal correlation might not be fully revealed.

\item \textbf{LightGBM~\cite{ke2017lightgbm}:} is a widely applied variant of the Gradient Boosted Machines Decision Tree (GBDT) classifier. To compute the information gain of possible split points and reduce the number of features, mutually exclusive features are bundled, and it samples data with large gradients, facilitating fast computation and high performance. We trained separate models for each context label and then combined the results. This is a non-graph-based method, which achieved good performance across many labels in prior works~\cite{gao2019human,ge2023heterogeneous}.

\item \textbf{ExtraMLP~\cite{vaizman2018context}:}  was a state-of-the-art  CA-HAR deep learning model, an MLP-based model that analyzed fine-grained handcrafted features extracted from both smartphone and smartwatch, and outperformed other models on the \textit{Extrasensory} dataset in \cite{vaizman2017recognizing}. Using a multi-label formulation, it can recognize multiple co-occurring context labels and mitigate imbalanced context labels.

\item \textbf{GCN~\cite{kipf2016semi}:} is a classic graph convolutional neural network that is essentially a simplified version of our proposed method. Two GCNs were stacked on the ordinary homogeneous graph in order to learn center node representations from 2-hop neighbors. It was included as a baseline because it previously demonstrated its ability to predict the purpose of a user's visit from geographic information (GPS-based mobility data)~\cite{martin2018graph}. 
Unlike {\model}, it considers all edges homogenous/same and does not consider the hyperedges or node/edge heterogeneity. Including it evaluated the utility of the hyperedges and heterogeneity properties of {\model}. 

\item \textbf{HHGNN~\cite{ge2023heterogeneous}:} built the same graph as we used in this paper. Node-heterogeneity and hypergraph properties were addressed by mapping different nodes with corresponding linear projection functions and applying Hypergraph Convolution Layers (HGC). However, a shared HGC was used for all types of hypergraphs, which overlooked and did not take maximal advantage of the edge heterogeneity property. Moreover, assigning specific linear functions for different kinds of nodes is an implicit and plain solution to the node-heterogeneity problem. In contrast, our {\model} addressed node-heterogeneity using explicit contrastive loss in the node embedding latent space.

\item \textbf{{\model}}: is our proposed framework. In contrast to the above-mentioned models, it not only addressed implicit node-heterogeneity using separate linear projections, but also explicitly addressed node-heterogeneity using a contrastive loss on nodes in embedding space. Furthermore, {\model} addressed edge-heterogeneity in a hypergraph by introducing separate hypergraph convolutional layers on different hyperedges and their connecting nodes.
\squishend

\subsection{Experimental Setup}

We randomly split each dataset into 60\% for training, 20\% for validation, and 20\% for hold-out testing. Grid search was used to determine optimal hyperparameter values for all models in our experiment. For our proposed model, the RAdam~\cite{liu2019variance} optimizer was utilized. The training batch size is fixed as 1024 and the epoch size is set to 300. We had learning rate of 8e-4 on the \textit{Extrasensory} dataset and 1e-3 on the \textit{WASH} dataset
The optimal values of $\lambda_1$ and $\lambda_2$ in the loss function were <0.03, 0.01> and <0.3, 0.1> for the \textit{WASH} and \textit{Extrasensory} datasets respectively.

\subsection{Evaluation Metrics}
\label{sec:metrics}
Due to the extremely imbalanced nature of the CA-HAR datasets, 
Matthews Correlation Coefficient (MCC), Eq.~\ref{eq:mcc}) and Macro F1 Score (MacF1, Eq.~\ref{eq:macf1}) were our main evaluation metrics. MCC is a statistical tool used for model evaluation. Its job is to gauge or measure the difference between the predicted values and actual values. In practice, MCC ranges from -1 to +1, and takes all elements in the confusion matrix (True Positive (TP), True Negative (TN), False Positive (FP), and False Negative (FN)) into account, making it a reliable statistical rate even for imbalanced datasets~\cite{chicco2020advantages}. F1-score is the harmonic mean of precision and recall. It is one of the most popular adopted evaluation metrics in classification tasks, and calculating the Macro F1 Score helped us to characterize the overall performance of the proposed model. 

\begin{equation}
MCC = \frac{TP \times TN - FP \times FN}{\sqrt{(TP+FP)(TP+FN)(TN+FP)(TN+FN)}}
\label{eq:mcc}
\end{equation}
\begin{equation}
    MacF1 = \frac{1}{C}\sum_{C_i=1}^{C}2*\frac{pre_{c_i}*rec_{c_i}}{pre_{c_i}+rec_{c_i}}
\label{eq:macf1}
\end{equation}
where C stands for the number of classes. ``pre'' and ``rec'' refer to precision and recall. Resultant values of metrics on specific labels are reported as well as averages of their category on the hold-out test set.

\begin{figure*}[t]
\centering

    \includegraphics[width=1.\linewidth]{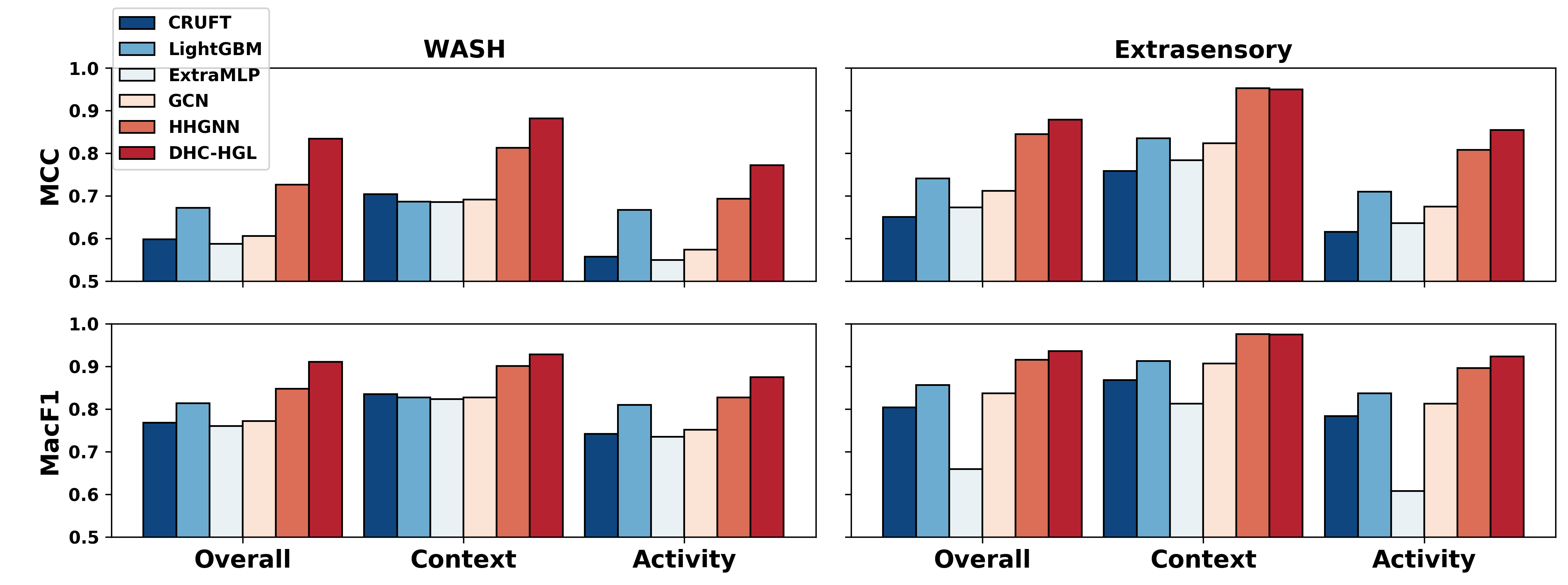}

    \caption{Average performance of all models across all labels (Overall) and on each category (Context, Activity). Our proposed {\model} is shown in red.}
    \label{fig:performance}

\end{figure*}

\begin{table}[t]
    \caption{Detailed Results on the  \textit{WASH} CA-HAR Dataset. For each label, the best results are marked in {\color{gray}gray}, and the second best results are \underline{underlined}. {\model} achieved the best performance on most labels.}

	\centering
    \small
\scalebox{.75}{
	\begin{tabular}{c||c|c c|c c| c c | c c | c c | c c c c}\hline
       \toprule
    \multirow{2}{*}{Category} &	\multirow{2}{*}{Label} & \multicolumn{2}{c|}{CRUFT}  & \multicolumn{2}{c|}{LightGBM} & \multicolumn{2}{c|}{ExtraMLP} & \multicolumn{2}{c|}{GCN} & \multicolumn{2}{c|}{HHGNN} & \multicolumn{4}{c}{\textbf{\model}} \\ 	
		& & MCC	& MacF1 & MCC & MacF1 & MCC	& MacF1 & MCC & MacF1 & MCC	& MacF1 & MCC & Impv. $\uparrow$ & MacF1 & Impv. $\uparrow$\\
        \midrule
\multirow{6}{*}{Context}	

        & In Pocket	            & 0.648	& 0.801 & 0.640  & 0.799  & 0.629  & 0.789  & 0.644  & 0.803 & \underline{0.771} & \underline{0.878} & \cellcolor{lightgray}0.849 & (13.3\%) & \cellcolor{lightgray}0.921 & (6.6\%)\\

	& In Hand 	            & 0.522	& 0.716 & 0.524  & 0.722  & 0.514  & 0.713  & 0.511  & 0.707 & \underline{0.718} & \underline{0.846} & \cellcolor{lightgray}0.781 & (11.1\%) & \cellcolor{lightgray}0.883 & (5.8\%)\\

	& In Bag	            & 0.756	& 0.867 & 0.706  & 0.837  & 0.683  & 0.821  & 0.701  & 0.833 & \underline{0.838} & \underline{0.914} & \cellcolor{lightgray}0.908 & (11.3\%) & \cellcolor{lightgray}0.953 & (5.7\%)\\

	& On Table-Face Down	& 0.803	& 0.896 & 0.802  & 0.896  & 0.805  & 0.897  & 0.819  & 0.904 & \underline{0.867} & \underline{0.931} & \cellcolor{lightgray}0.942 & (11.1\%)  & \cellcolor{lightgray}0.971 & (5.5\%)\\

	& On Table-Face Up 	    & 0.793 & 0.894 & 0.862  & 0.880  & 0.800  & 0.898  & 0.781  & 0.890 & \underline{0.872} & \underline{0.936} & \cellcolor{lightgray}0.928 & (8.1\%) & \cellcolor{lightgray}0.964 & (3.8\%)\\ 
        \cmidrule{2-16}
    &\textbf{Context Avg} &0.704 &   0.835   &   0.687   &0.827  &0.686  &0.823  &0.691& 0.827   &\underline{0.813}  &   \underline{0.901}   &\cellcolor{lightgray}0.902 &    (10.9\%)  &\cellcolor{lightgray}0.950 & (5.4\%)\\
        \midrule
\multirow{14}{*}{Activity}	
        & Lying Down 	        & 0.820 & 0.907  & 0.783	& 0.887 & 0.831  & 0.912  & 0.848  & 0.921  & \underline{0.893}	& \underline{0.945}  & \cellcolor{lightgray}0.943 & (8.3\%)  & \cellcolor{lightgray}0.971 & (4.0\%)\\

	& Sitting 	            & 0.758 & 0.876  & 0.735	& 0.864 & 0.741  & 0.867  & 0.743  & 0.686  & \underline{0.816}	& \underline{0.907}  & \cellcolor{lightgray}0.898 & (12.7\%)  & \cellcolor{lightgray}0.949 & (5.8\%)\\

	& Walking	            & 0.510 & 0.721  & 0.536	& 0.741 & 0.523  & 0.731  & 0.519  & 0.732  & \underline{0.623}	& \underline{0.794}  & \cellcolor{lightgray}0.738 & (23.9\%)  & \cellcolor{lightgray}0.860 & (10.7\%)\\

	& Sleeping	            & 0.915 & 0.957  & 0.922	& 0.961 & 0.921  & 0.960  & 0.935  & 0.967  & \underline{0.954}	& \underline{0.977}  & \cellcolor{lightgray}0.947 & (3.1\%)  & \cellcolor{lightgray}0.987 & (1.5\%)\\

	& Talking On Phone	    & 0.417 & 0.645  & 0.545	& 0.730 & 0.436  & 0.656  & 0.465  & 0.675  & \underline{0.657}	& \underline{0.806}  & \cellcolor{lightgray}0.760 & (12.9\%)  & \cellcolor{lightgray}0.869 & (6.5\%)\\

	& Bathroom 	            & 0.425 & 0.660  & 0.499	& 0.704 & 0.427  & 0.659  & 0.406  & 0.641  & \underline{0.615}	& \underline{0.780}  & \cellcolor{lightgray}0.718 & (19.8\%)  & \cellcolor{lightgray}0.846 & (10.0\%)\\

	& Standing	            & 0.457 & 0.681  & 0.486	& 0.703 & 0.463  & 0.686  & 0.489  & 0.705  & \underline{0.605}	& \underline{0.777}  & \cellcolor{lightgray}0.716 & (27.4\%)  & \cellcolor{lightgray}0.845 & (12.9\%)\\

	& Jogging	            & 0.551 & 0.737  & \cellcolor{lightgray}0.964	& \cellcolor{lightgray}0.982 & 0.520  & 0.712  & 0.599  & 0.765  & 0.712	& 0.837  & \underline{0.717} & (-11.2\%)  & \underline{0.841} & (-5.9\%)\\

	& Running 	            & 0.479 & 0.692  & \cellcolor{lightgray}0.946	& \cellcolor{lightgray}0.973 & 0.420  & 0.647  & 0.489  & 0.695  & 0.604	& 0.769  & \underline{0.680} & (-22.7\%)  & \underline{0.819} & (-12.5\%)\\

	& Stairs-Going Down	    & 0.382 & 0.636   & 0.488	& 0.697 & 0.376  & 0.626  & 0.374  & 0.624  & \underline{0.529}	& \underline{0.726}  & \cellcolor{lightgray}0.623 & (23.3\%)  & \cellcolor{lightgray}0.789 & (11.1\%)\\

	& Stairs-Going Up	    & 0.397 & 0.645  & 0.469	& 0.686 & 0.388  & 0.634  & 0.399  & 0.640  & \underline{0.537}	& \underline{0.731}  & \cellcolor{lightgray}0.638 & (27.7\%)  & \cellcolor{lightgray}0.799 & (13.4\%)\\

	& Typing	            & 0.636 & 0.793  & 0.666	& 0.814 & 0.636  & 0.794  & 0.672  & 0.817  & \underline{0.770}	& \underline{0.876}  & \cellcolor{lightgray}0.865 & (16.2\%)  & \cellcolor{lightgray}0.930 & (8.0\%)\\

	& Exercising	        & 0.493 & 0.699  & 0.626	& 0.783 & 0.463  & 0.672  & 0.523  & 0.715  & \underline{0.692}	& \underline{0.826}  & \cellcolor{lightgray}0.769 & (14.9\%)  & \cellcolor{lightgray}0.874 & (7.7\%)\\
    \cmidrule{2-16}
    &\textbf{Activity Avg} &0.557    & 0.742 & 0.667 &   0.810   &0.550  &   0.735   &0.574  &0.751  &\underline{0.693}  &\underline{0.827}  &\cellcolor{lightgray}0.808 &    (16.7\%) &\cellcolor{lightgray}0.897  &   (8.4\%)\\
    
    \midrule
    \multicolumn{2}{c|}{\textbf{Overall Avg}} & 0.598    &0.768  &0.672  &0.814  &0.588  &0.760  &0.606  &0.772  &\underline{0.726}  &\underline{0.848}  &\cellcolor{lightgray}0.834 &   (14.9\%) &\cellcolor{lightgray}0.911&   (7.5\%)\\
        \bottomrule
	\end{tabular}}

	\label{tab:detailExpResultUnscripted}
\end{table}

\subsection{Experiment Results}
\label{sec:results}

We show the overall and average performance of all models on each category in Fig.~\ref{fig:performance}. Detailed results for specific labels on two real-world CA-HAR datasets are reported in Tables~\ref{tab:detailExpResultUnscripted} and ~\ref{tab:detailExpResultExtrasensory}. Our findings derived from these experimental results are described in this section.

\smallskip\noindent\textbf{Overall performance:} As shown in Fig.~\ref{fig:performance}, {\model} consistently achieves significantly better recognition over baseline models across different datasets, highlighting the effectiveness and generalizability of our proposed model. The success of both HHGNN and {\model} demonstrates that explicitly encoding the user and context in the CA-HAR graph can capture correlations between them and facilitate learning better representations. Based on the same initial heterogeneous hypergraph, the better performance of \model~can prove that our deep heterogeneity design further benefits CA-HAR tasks. To be specific, \model~outperform the best baseline, HHGNN, with an overall improvement of 14.9\% and 4.1\% in MCC and 7.5\% and 2.2\% in Macro F1 on the WASH and ExtraSensory datasets, respectively. 

\smallskip\noindent\textbf{Model performance on each label category:} On the \textit{WASH} dataset, MCC and Macro F1 improved by 10.9\% and 5.4\% for context (phone placement) and improved by 16.7\% and 8.4\% for activity recognition. On the \textit{Extrasensory} dataset, MCC and Macro F1 results improved by 5.8\% and 3.0\% for activity recognition and had performance on par with the best baseline for context category (i.e., the performance difference is less than 0.3\%). As most CA-HAR tasks consider activities as the most important category to recognize, the observation of higher performance gains in the activity category is encouraging.

\smallskip\noindent\textbf{Detailed results for each specific label on the \textit{WASH} CA-HAR dataset:}
Detailed results for all models on the \textit{WASH} dataset are listed in Table~\ref{tab:detailExpResultUnscripted}. 
Among the baseline models, first, it can be observed that HHGNN outperformed all other models, which strongly demonstrates the advantage of modeling CA-HAR tasks as a heterogeneous hypergraph. Additionally, {\model} outperforms all baselines on most labels by a large margin, except for the \textit{Jogging} and \textit{Running} labels, which have very small positive support instances (are scarce) (with only \num[group-separator={,}]{3804} and \num[group-separator={,}]{2422} positive instances, separately) compared to other labels with larger positive instance support. For instance, \textit{Sleeping} has \num[group-separator={,}]{2935264} positive instances.
On the other hand, \textit{Jogging} and \textit{Running} produce similar sensor signals, making them challenging to discriminate~\cite{chakraborty2008view}. LightGBM overcomes this confusion by training a separate model for each label. Other baselines that train a single model for all labels achieved are less able to discriminate \textit{Jogging} and \textit{Running}. Another possible explanation is that as labels were self-reported, many participants may have confused \textit{Jogging} and \textit{Running}. It is instructive to note that when compared to the best overall baseline HHGNN, {\model} managed to improve on the \textit{Jogging} and \textit{Running} labels.

\begin{table}[t]
    \caption{Detailed Results on the \textit{Extrasensory} CA-HAR Dataset. For each label, the best results are in {\color{gray}gray}, and the second best results are \underline{underlined}. {\model} achieved the best performance on all activities and performance mostly on-par with or better than the best baselines.}

	\centering
    \small
    \scalebox{.75}{
	\begin{tabular}{c||c|c c|c c| c c | c c | c c | c c c c}\hline
       \toprule
    \multirow{2}{*}{Category} &	\multirow{2}{*}{Label} & \multicolumn{2}{c|}{CRUFT}  & \multicolumn{2}{c|}{LightGBM} & \multicolumn{2}{c|}{ExtraMLP} & \multicolumn{2}{c|}{GCN} & \multicolumn{2}{c|}{HHGNN} & \multicolumn{4}{c}{\textbf{\model}} \\ 	
		& & MCC	& MacF1 & MCC & MacF1 & MCC	& MacF1 & MCC & MacF1 & MCC	& MacF1 & MCC & Impv. $\uparrow$ & MacF1 & Impv. $\uparrow$\\
        \midrule
\multirow{5}{*}{Context}	
        & In Pocket  	& 0.782 & 0.884  & 0.839	& 0.916 & 0.801   & 0.825  & 0.809  & 0.898  & \underline{0.950} & \underline{0.974} & \cellcolor{lightgray}0.952 & (0.2\%)  & \cellcolor{lightgray}0.976 & (0.1\%)\\

	& In Hand 	    & 0.638 & 0.797  & 0.736	& 0.855 & 0.682   & 0.686  & 0.747  & 0.864   &\cellcolor{lightgray} 0.931 & \cellcolor{lightgray}0.965 & \underline{0.925}  & (-0.7\%) & \underline{0.962} & (-0.4\%)\\

	& In Bag 	    & 0.775 & 0.878  & 0.895	& 0.945 & 0.788   & 0.786  & 0.862  & 0.928   & \cellcolor{lightgray}0.964	& \cellcolor{lightgray}0.982 & \underline{0.954} & (-1.1\%)  & \underline{0.976} & (-0.6\%)\\

	& On Table	    & 0.835 & 0.915  & 0.872	& 0.935 & 0.864   & 0.956  & 0.876  & 0.937   & \underline{0.966}	& \underline{0.983} & \cellcolor{lightgray}0.969  & (0.3\%) & \cellcolor{lightgray}0.984 & (0.1\%)\\
        \cmidrule{2-16}
    &\textbf{Context Avg}    & 0.758    &   0.868   &   0.835   &   0.913   &   0.784   &0.813  &   0.824   &  0.907  & \cellcolor{lightgray}0.953   &    \cellcolor{lightgray}0.976   & \underline{0.950}  & (-0.3\%) & \underline{0.975} & (-0.2\%)\\
        \midrule

\multirow{13}{*}{Activity}	
        & Lying Down	       & 0.918 & 0.959  & 0.911	& 0.955 & 0.936   & 0.958  & 0.941  & 0.971   & \underline{0.962}	& \underline{0.981} & \cellcolor{lightgray}0.977 & (1.6\%)  & \cellcolor{lightgray}0.989 & (0.8\%)\\

	& Sitting	           & 0.772 & 0.885  & 0.755	& 0.876 & 0.797   & 0.889  & 0.781  & 0.888   & \underline{0.870}	& \underline{0.935} & \cellcolor{lightgray}0.907  & (4.3\%) & \cellcolor{lightgray}0.953 & (2.0\%)\\

	& Walking  	           & 0.535 & 0.730  & 0.560	& 0.746 & 0.554   & 0.539  & 0.580  & 0.758   & \underline{0.718}	& \underline{0.846} & \cellcolor{lightgray}0.772 & (7.5\%)  & \cellcolor{lightgray}0.878 & (3.8\%)\\

	& Sleeping	           & 0.934 & 0.967  & 0.934	& 0.967 & 0.950   & 0.965  & 0.952  & 0.976   & \underline{0.979}	& \underline{0.989} & \cellcolor{lightgray}0.985 & (0.6\%)  & \cellcolor{lightgray}0.993 & (0.3\%)\\

	& Talking	           & 0.635 & 0.794  & 0.697	& 0.833 & 0.681   & 0.703  & 0.740  & 0.859   & \underline{0.858}	& \underline{0.927} & \cellcolor{lightgray}0.895  & (4.3\%) & \cellcolor{lightgray}0.946 & (2.1\%)\\

	& Bath-Shower	       & 0.495 & 0.703  & 0.729	& 0.848 & 0.496   & 0.404  & 0.648  & 0.798   & \underline{0.780}   & \underline{0.880} & \cellcolor{lightgray}0.827  & (5.9\%) & \cellcolor{lightgray}0.908 & (3.1\%)\\

	& Toilet               & 0.416 & 0.661  & 0.524 & 0.716 & 0.429   & 0.326  & 0.446  & 0.666   & \underline{0.670}   & \underline{0.813} & \cellcolor{lightgray}0.743 & (10.8\%)  & \cellcolor{lightgray}0.859 & (5.7\%)\\
	
	& Standing	           & 0.579 & 0.762  & 0.573	& 0.760 & 0.621   & 0.648  & 0.639  & 0.802   & \underline{0.766}	& \underline{0.875} & \cellcolor{lightgray}0.828  & (8.1\%) & \cellcolor{lightgray}0.911 & (4.0\%)\\

	& Running	           & 0.609 & 0.782  & 0.656	& 0.802 & 0.616   & 0.555  & 0.724  & 0.845   & \underline{0.866}	& \underline{0.930} & \cellcolor{lightgray}0.883  & (1.9\%) & \cellcolor{lightgray}0.938 & (0.9\%)\\

	& Stairs-Going Down    & 0.431 & 0.674  & \underline{0.791}	& \underline{0.886} & 0.441   & 0.334  & 0.462  & 0.676   & 0.695	& 0.831 & \cellcolor{lightgray}0.807 & (2.1\%)  & \cellcolor{lightgray}0.898 & (1.3\%)\\

	& Stairs-Going Up	   & 0.429 & 0.674  & 0.654	& 0.801 & 0.446   & 0.341  & 0.513  & 0.710   & \underline{0.706}	& \underline{0.837} & \cellcolor{lightgray}0.770  & (9.1\%) & \cellcolor{lightgray}0.876 & (4.6\%)\\

	& Exercising	       & 0.642 & 0.800  & 0.735	& 0.852 & 0.665   & 0.636 & 0.673  & 0.812   & \underline{0.803}    & \underline{0.909} & \cellcolor{lightgray}0.868  & (4.5\%) & \cellcolor{lightgray}0.930 & (2.3\%)\\ 
    \cmidrule{2-16}
    &\textbf{Activity Avg} &    0.616&0.783 &   0.710   &   0.837   &0.636  &0.608  &0.675  &0.813  & \underline{0.808}  & \underline{0.896}  &\cellcolor{lightgray}0.855 & (5.8\%) &\cellcolor{lightgray}0.923& (3.0\%)\\
    \midrule
    \multicolumn{2}{c|}{\textbf{Overall Avg}} &  0.651&  0.804   &0.741  &0.856  &0.673  &0.659  &0.712  &0.837  & \underline{0.845}  & \underline{0.916}  &\cellcolor{lightgray}0.879& (4.1\%)  &\cellcolor{lightgray}0.936& (2.2\%)\\
 \bottomrule
	\end{tabular}}

	\label{tab:detailExpResultExtrasensory}
\end{table}

\smallskip\noindent\textbf{Detailed results for each specific label on the \textit{Extrasensory} CA-HAR dataset:}
Detailed experimental results of all models for all labels are reported in Table~\ref{tab:detailExpResultExtrasensory}. It can be observed that slightly different from \textit{WASH} dataset, {\model} outperforms baselines on all activity labels, even including labels where HHGNN could not outperform LightGBM (e.g., Stairs-Going Down). Meanwhile, {\model} did not achieve as significant improvements on context labels on \textit{Extrasensory} as it did on the \textit{WASH} dataset. It is still worth noting that results for {\model} on context labels are on-par with the best baselines, with a difference of no more than 1.1\%. The lack of improvement on context labels might be due to the limited amount of $\{u,c\}$ edges in \textit{WASH} as reported in Fig.~\ref{fig:edgecount}, where only 1.2\% of total hyperedges are $\{u,c\}$ in \textit{Extrasensory}, compared to 9.6\% in \textit{WASH}. 

Nevertheless, most of the results of {\model} on the \textit{Extrasensory} dataset are consistent with results achieved on the \textit{WASH} dataset.

\begin{figure*}[ht]
\centering

    \includegraphics[width=.9\linewidth]{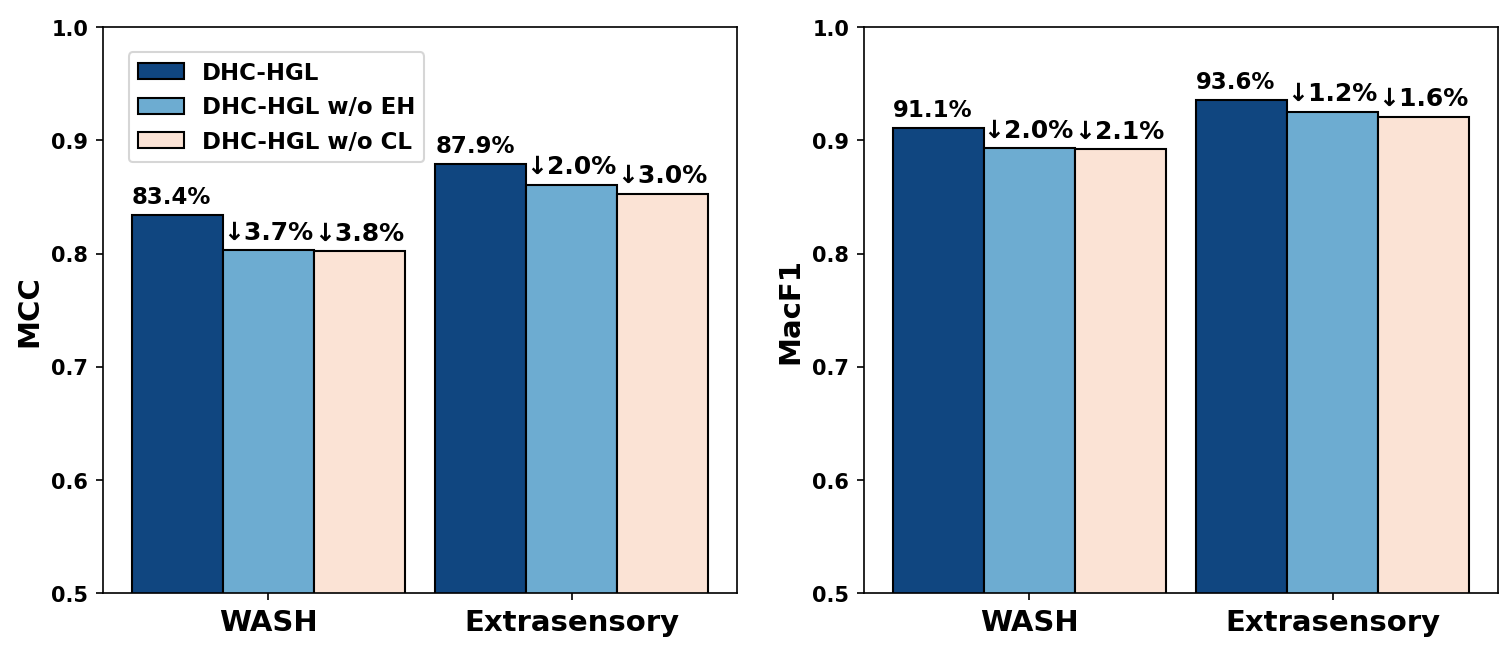}

    \caption{Ablation study compares {\model} with its variants. EH: designs addressing edge-heterogeneity. CL: designs introducing contrastive loss for node-heterogeneity.}
    \label{fig:performanceAB}

\end{figure*}
\begin{figure}[htbp]
    \centering
    \subfloat[Original nodes]{
        \includegraphics[width=0.232\linewidth]{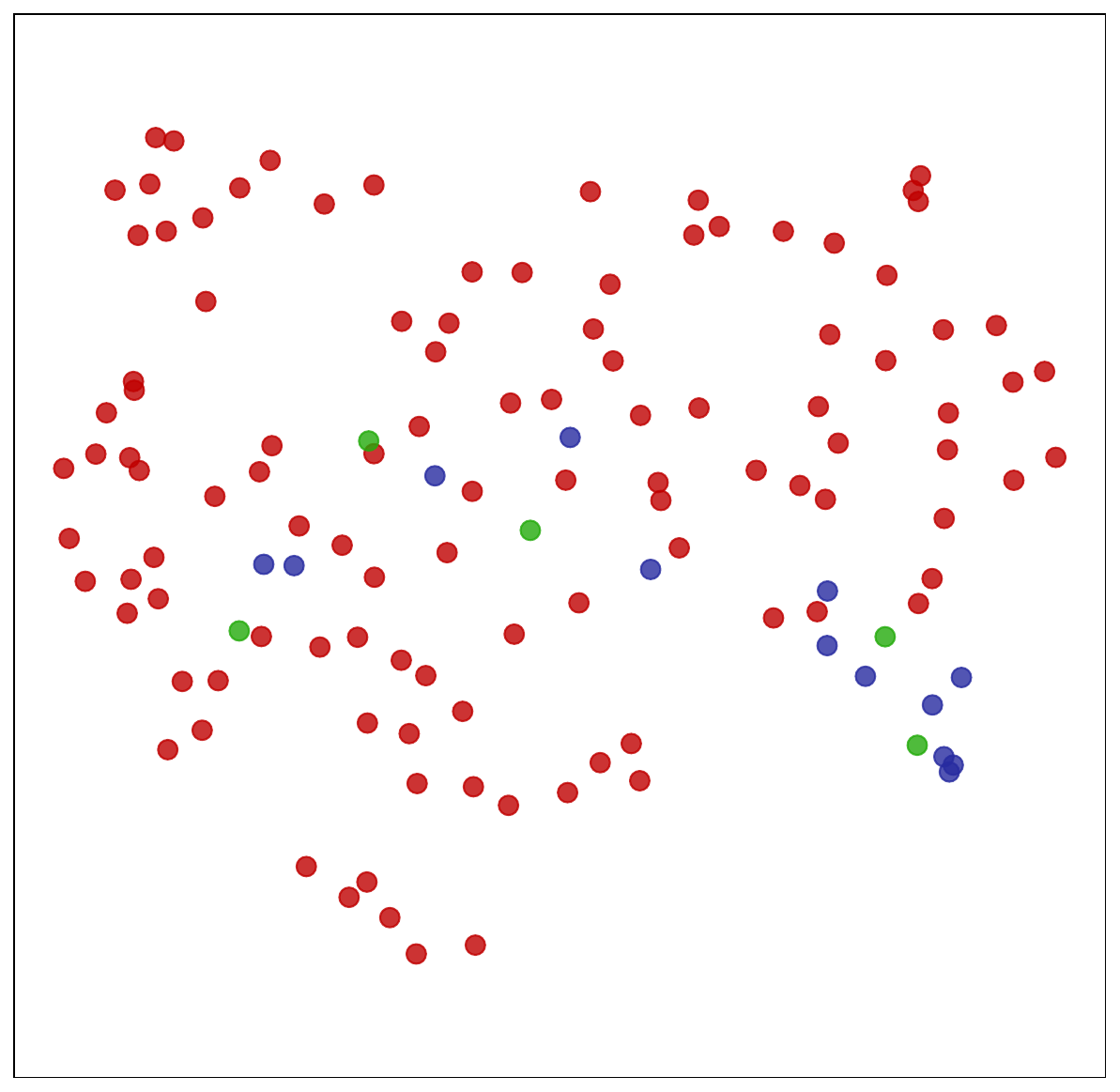}
        \label{fig:umapOriginal}
    } 
    \subfloat[HHGNN learned nodes]{
        \includegraphics[width=0.232\linewidth]{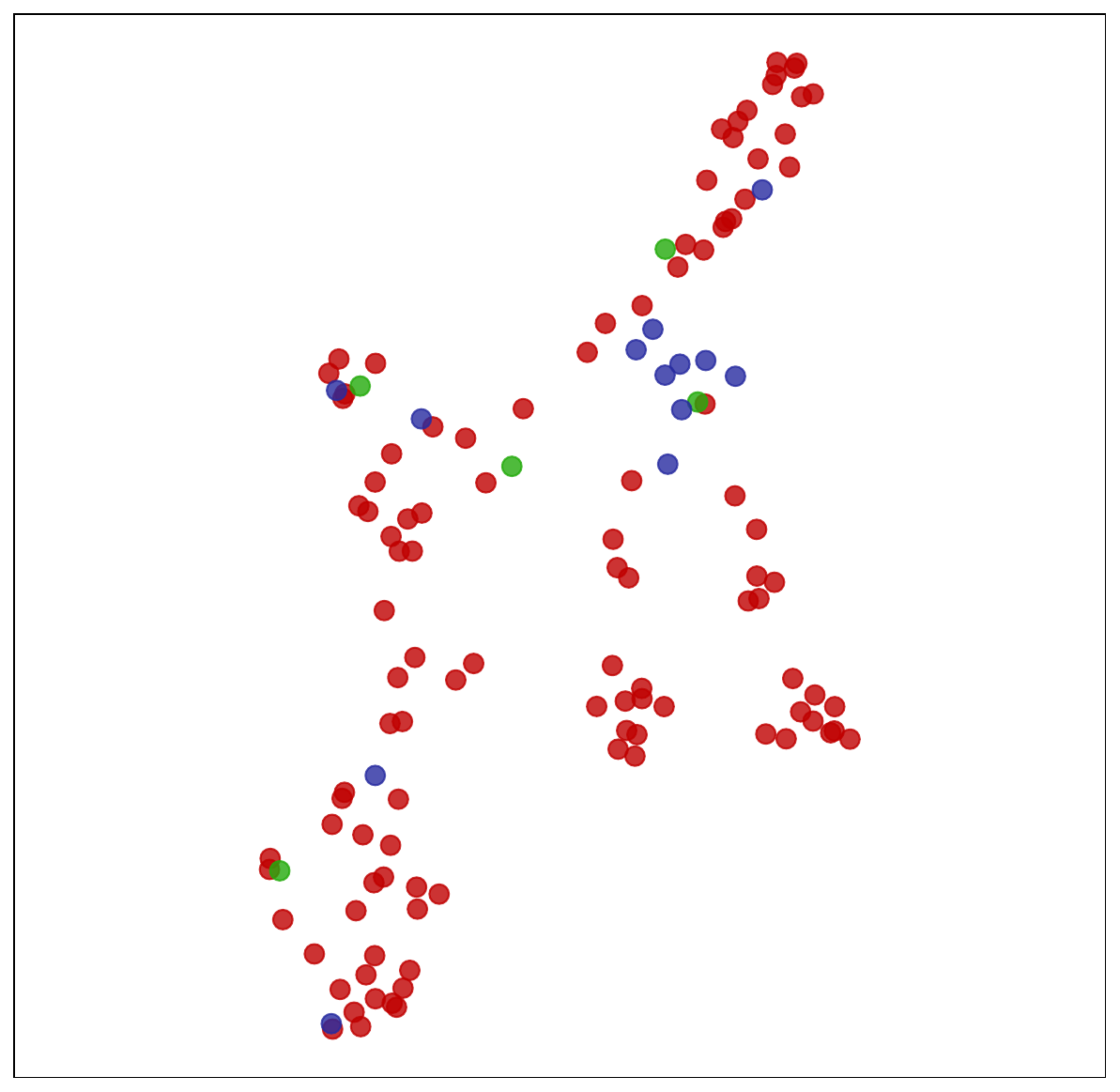}
        \label{fig:umapHHGNN}
    } 
    \subfloat[\model~w/o CL]{
        \includegraphics[width=0.232\linewidth]{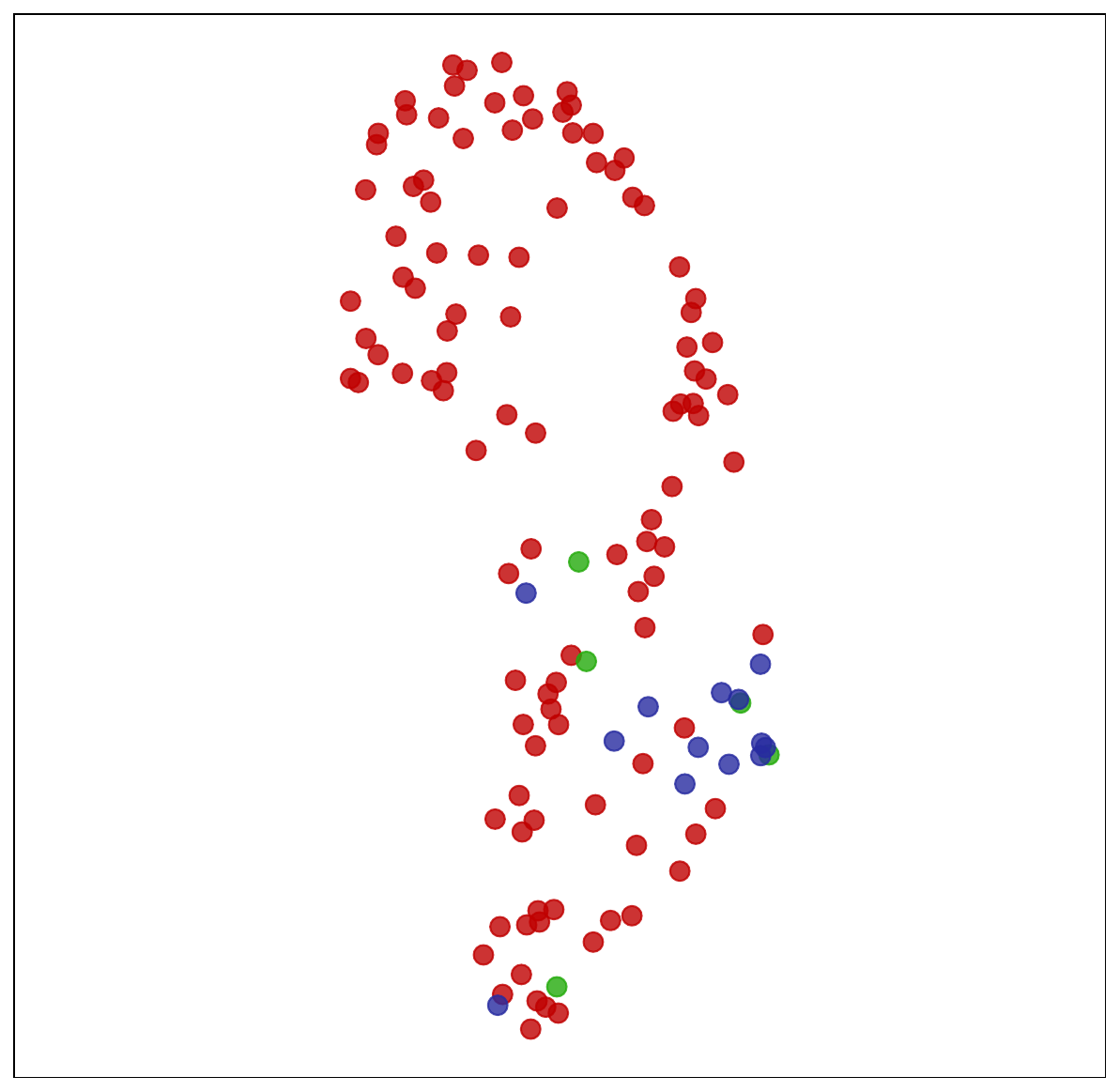}
        \label{fig:umapLearnedwoCL}
    }
    \subfloat[\model~w/ CL]{
        \includegraphics[width=0.232\linewidth]{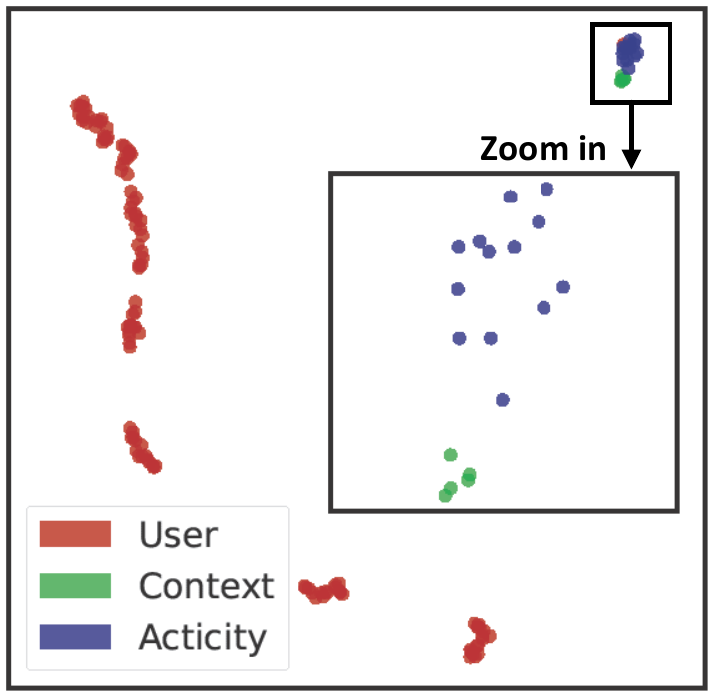}
        \label{fig:umapLearnedwCL}
    }
    \caption{UMAP visualization of \ref{fig:umapOriginal}) the original \textit{WASH} initialized node, \ref{fig:umapHHGNN}) HHGNN learned node embedding, \ref{fig:umapLearnedwoCL}) node embedding learned by \model~without contrastive loss, and \ref{fig:umapLearnedwCL}) node embedding learned by \model~with contrastive loss (full model). The red, green, and blue dots in the graph represent user, context, and activity nodes, respectively.}

    \label{fig:umap}
\end{figure}

\section{Analysis}
\label{sec:analysis}
In this section, we attempt to dive deeper into our results and derive additional insights on the performance of our proposed {\model} in a Q\&A fashion.

\smallskip\noindent\textbf{RQ1: How much does each proposed novel component contribute?} 
In order to deal with the heterogeneous nature of CA-HAR data effectively, our proposed {\model} framework integrates several key components. A key question that arises is the utility of the various components of {\model}. To answer this question, an ablationt study was conducted on both CA-HAR datasets in order to understand the contributions made by the contrastive loss and edge heterogeneity aspects of the {\model} design to its improved performance for the CA-HAR task. Results are shown in Fig.~\ref{fig:performanceAB}. The models explored are described below:
\begin{itemize}
    \item \textbf{\model}: Proposed approach using optimal hyperparameter values.
    \item \textbf{\model~w/o edge heterogeneity(EH)}: The hyperedge splitting procedure was removed, which yielded an integrated graph that is passed through a common, single hyperConv layer. 
    \item \textbf{{\model~w/o contrastive loss (CL)}}: {The contrastive loss was removed from \model. In this way, the graph learned is only used for classification. It is included in order to evaluate the contribution of the contrastive loss to both model performance and interpretability aspects.}
\end{itemize}
Unsurprisingly,  performance in terms of MCC/MacF1  decreased by -3.7\%/-2.0\% and -2.0\%/-1.2\% when the contrastive loss was removed; and decreased by -3.8\%/-2.1\% and -3.0\%/-1.6\% when the same hypergraph convolution layer was used without distinguishing the type of hyperedges that occur in the \textit{WASH} and \textit{Extrasensory} datasets, respectively. In summary, 1) both the innovative contrastive loss and passing different types of hypergraphs through different hypergraph convolution layers had non-trivial contributions to the overall performance improvement of {\model}, and 2) using different hypergraph convolutional layers to address edge-heterogeneity appeared to have a larger influence than the designed contrastive loss. This is understandable as no other component addresses edge-heterogeneity. In contrast, the contrastive loss function addressed node-heterogeneity, which is partly resolved by separate linear projections as proposed in HHGNN and leveraged in {\model}.

\smallskip\noindent\textbf{RQ2: Does {\model} accurately capture the relationships between various nodes?}
We previously suggested that {\model} captured predictive relationships between various entities within context labels. 
In order to validate this claim, here, in Fig.~\ref{fig:umap}, we present UMAP~\cite{mcinnes2018umap} visualizations of node representations generated from the \textit{WASH} dataset when various graph-based models were applied. Visualizations on the \textit{Extrasensory} dataset followed a similar pattern.

UMAP projects representations from high-dimensional to two-dimensional space while preserving both local properties (within each cluster) and global properties (among clusters). Essentially, nodes that are close in high dimensions remain close after UMAP projection. 
The visualization on the left (Fig~\ref{fig:umapOriginal}) shows that the initialized points are close to being uniformly distributed. HHGNN tried to deal with node heterogeneity by assigning specific linear functions to different types of nodes, but the learned node embedding (Fig.~\ref{fig:umapHHGNN}) is not able to distinguish different types of nodes well, which is consistent with our previous claim that separate linear functions are not adequate to handle node heterogeneity. By applying {\model}, nodes form clusters with clear boundaries (Fig.~\ref{fig:umapLearnedwCL}). Additionally, user nodes formed different groups, suggesting the possibility of building customized models.
{A comparison of Figs~\ref{fig:umapLearnedwoCL} and ~\ref{fig:umapLearnedwCL} further demonstrates the contribution of contrastive loss not only quantitatively but also its improvement of interpretability.}

\begin{figure}[htbp]
    \centering
    \subfloat[The overall performance (MCC) of {\model} with different node embedding sizes on the \textit{WASH} dataset.]{
        \includegraphics[width=0.45\linewidth]{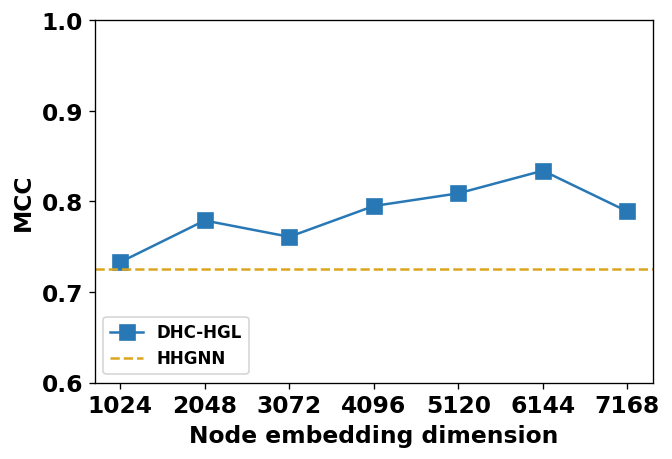}
        \label{fig:commonDmccwash}
    } 
    \hspace{0.5cm}
    \subfloat[The overall performance (MCC) of {\model} with different node embedding size on the\textit{Extrasensory} dataset.]{
        \includegraphics[width=0.45\linewidth]{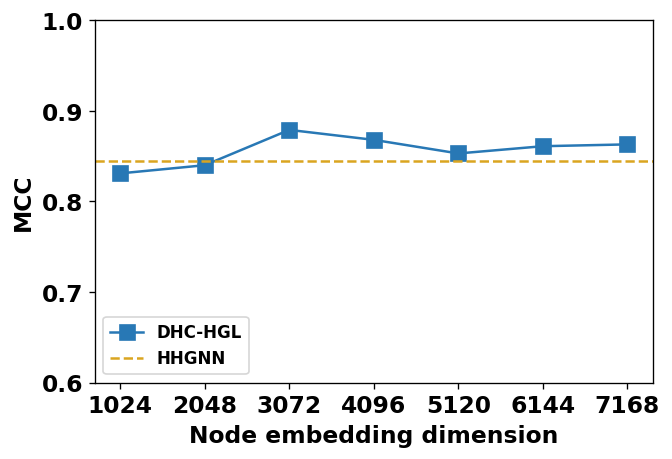}
        \label{fig:commonDmccextra}
    } 
    \
    \subfloat[The overall performance (MCC) of {\model} with different numbers of graph learning layers on the \textit{WASH} dataset.]{
        \includegraphics[width=0.45\linewidth]{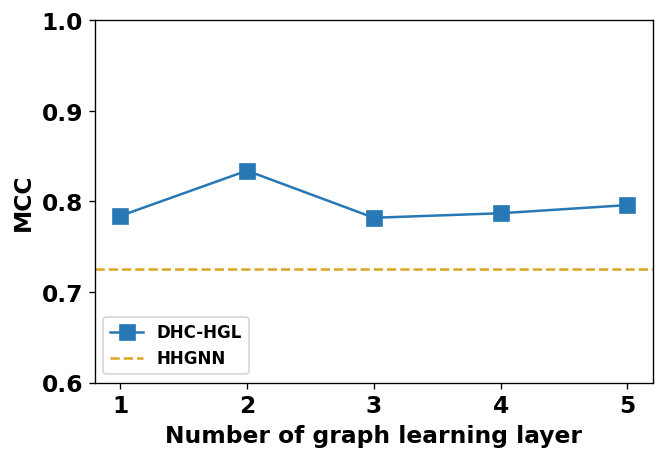}
        \label{fig:layermccwash}
    } 
    \hspace{0.5cm}
    \subfloat[The overall performance (MCC) of {\model} with different numbers of graph learning layers on the \textit{Extrasensory} dataset.]{
        \includegraphics[width=0.45\linewidth]{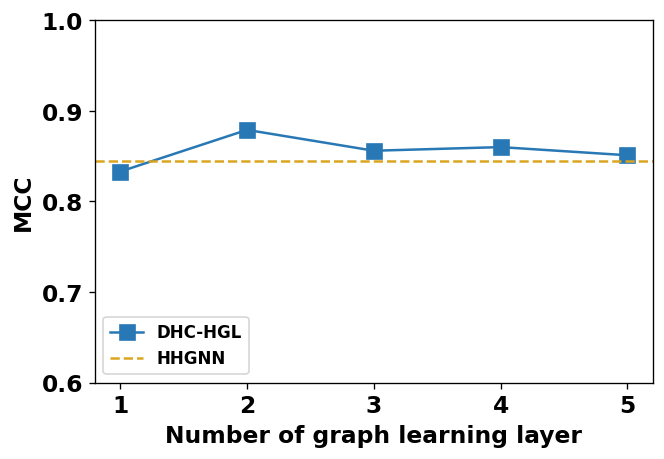}
        \label{fig:layermccextra}
    }

    \caption{Hyperparameter evaluation.}

    \label{fig:hypers}
\end{figure}

\smallskip\noindent\textbf{RQ3: How does model structure influence recognition performance?}
We explore the robustness of {\model} by investigating the impact of two key hyperparameters: 1) node embedding size and 2) the number of graph learning layers. A performance visualization is shown in Fig.~\ref{fig:hypers} with four subplots. The best-performing baseline HHGNN with a node embedding dimension of 1536\footnote{These are the best hyperparameters found through grid search.} is included as a yellow line in all subplots. 

\smallskip\noindent\textit{Impact of node embedding dimension in Fig.~\ref{fig:commonDmccwash} and Fig.~\ref{fig:commonDmccextra}: }

\begin{enumerate}
    \item {\model} achieved its best performance on the \textit{WASH} dataset with 6144 dimensions and on the \textit{Extrasensory} dataset with 3072 dimensions. Overall, a performance gain can be observed as the embedding dimension increases, where a higher dimension enables the model to capture more information from the graph. However, there is also a saturation point at which the performance stops increasing after the embedding dimension is larger than the optimal value.
    \item {\model} consistently outperformed HHGNN in most cases in the regions of the graphs around the optimal node embedding dimensions, which indicates its robustness. A larger gain is achieved on the \textit{WASH} dataset compared to the \textit{Extrasensory} dataset.
\end{enumerate}
\noindent\textit{Impact of number of graph learning layer in Fig.~\ref{fig:layermccwash} and Fig.~\ref{fig:layermccextra}:}
\begin{enumerate}
    \item {\model} achieved its best performance with two graph learning layers, on both \textit{WASH} and \textit{Extrasensory} datasets. The performance was reduced when more than two graph learning layers were used. One possible reason is that deeper graph learning layers make it more difficult for the model to generalize and are also prone to overfitting. Another possible reason is that deeper graph learning layers might cause over-smoothing~\cite{li2018deeper,zhou2020graph}, where learned node embeddings become similar and indistinguishable. 
    \item {\model} yields better performance than HHGNN under most cases, indicating {\model}'s robustness w.r.t the number of graph learning layers.
\end{enumerate}

\section{Limitations}
Our work has a few limitations we now mention. First, our definition of context as phone placement in the CA-HAR problem statement followed and was inspired by prior work. Interested researchers may extend our work to other context definitions subject to availability of data on other relevant contextual information such as device type. Secondly, while we formed the heterogeneous hypergraph on the training set in a user-aware mode, the inference on the test set was user-agnostic / user-implicit (i.e., we did not use user identity as an input while inferencing, nor did we infer user nodes given input sensor signal). We leave user-explicit inference and inferring user identity as future work as it requires minimum adaptation effort. Admittedly, we evaluated the effectiveness of our framework mainly on the hypergraph convolutional neural network backbone. We believe that our design is backbone-agnostic and that our novel contributions can serve as plug-ins for various hypergraph learning backbones. Lastly, casting the CA-HAR task into a graph learning task is only one of the possible approaches for resolving the problem. We may consider further combining other modalities with the message-passing design to improve the model performance.

\section{Conclusion and Future work}
\label{sec:conclusion}

Context-aware Human Activity Recognition is an emerging but challenging problem for academia and industry domains. Prior work mainly researched non-graph-based, feature-dependent ordinary-graph-based, or hypergraph-based methods and models that implicitly addressed node-heterogeneity. In this work, we introduced a novel feature-independent hypergraph-based neural networks approach {\model}, which addresses both edge-heterogeneity and node-heterogeneity in a CA-HAR data-transformed hypergraph. More specifically, 1) for handling edge-heterogeneity, we leverage different hypergraph convolutional layers for various edge types in contrast to unified hypergraph convolutional layers with shared parameters for all hyperedges. 2) To enforce explicit node-heterogeneity, we designed a contrastive loss applied to the node embedding latent space. Such a contrastive loss enlarges the distance between different types of nodes while pulling nodes of the same type closer. In an extensive experimental study,  {\model} yielded a superior performance boost on two large datasets and provided better UMAP visualizations of label representation distributions, which enhances model interpretability. 

In future work, we plan to evaluate our design on more neural network backbones other than GCN-based models as our model is intuitively backbone-agnostic. We adopted the straightforward summation function as our aggregation method of hypergraph neural networks. In the future, we will explore other possibilities, such as mean pooling and attention-based pooling. Following prior works, our definition of context was restricted to the device placement in this work due to the limited availability of other contextual factors (e.g., device type) in experimental CA-HAR datasets. However, in the future, it would be interesting to explore whether exploiting other context labels/features can help to further improve the model performance. 
In any case, our novel designed components can easily be adapted to incorporate other contextual factors with minimum effort. {We would also like to point out the possibility of extending this work to handle both user-explicit and user-implicit CA-HAR tasks by introducing user identity as part of the input. This will facilitate user-explicit settings, introducing unknown user nodes for user-implicit settings, or inferring not only contexts and activities but also user identities. Although this work focused on CA-HAR tasks, {\model} also has the potential applicability to other general time series problems and link prediction problems, including diagnosing cardiovascular disease (CVD) using Electrocardiogram (ECG) signals and document recommendation.}

\section*{Acknowledgments}
The research was sponsored by DARPA grant HR00111780032-WASH-FP-031. We thank the WPI WASH team for gathering the WASH dataset analyzed in this paper.
\bibliographystyle{ACM-Reference-Format}
\bibliography{sample-base}










\end{document}